\documentclass[11pt]{article}

\usepackage[final]{acl}

\usepackage{times}
\usepackage{latexsym}

\usepackage{worldflags}

\definecolor{mygray}{HTML}{EBEEEA}

\usepackage[T1]{fontenc}

\usepackage[utf8]{inputenc}

\usepackage{microtype}

\usepackage{inconsolata}

\usepackage{graphicx}

\usepackage{booktabs}
\usepackage{subcaption}
\usepackage{adjustbox}
\usepackage{threeparttable}
\usepackage{multicol,multirow}
\usepackage[skins]{tcolorbox}

\usepackage{pifont}
\newcommand{\cmark}{\ding{51}}%
\newcommand{\xmark}{\ding{55}}%
\usepackage{arydshln}
\usepackage{amsmath}
\usepackage{amsfonts}
\usepackage{tikz}
\usepackage{xcolor}
\usepackage{colortbl}

\definecolor{customblue}{RGB}{100,111,250}
\definecolor{customred}{RGB}{239,85,59}
\definecolor{customblue2}{RGB}{102,102,204}
\definecolor{mamiColor}{RGB}{206,39,39}
\definecolor{existColor}{RGB}{141,48,228}

\usepackage{enumitem}

\usepackage{pgfplots}
\usepackage{pgfplotstable}
\usepgfplotslibrary{groupplots}
\usepgfplotslibrary{statistics}
\usetikzlibrary{matrix} 
\pgfplotsset{compat=newest}
\usetikzlibrary{pgfplots.fillbetween,patterns} 

\title{\textsc{MemeWeaver}: Inter-Meme Graph Reasoning for\\ Sexism and Misogyny Detection}

\newcommand{\samethanks}[1][\value{footnote}]{\footnotemark[#1]}

\author{
  \textbf{Paolo Italiani\thanks{Equal contribution (co-first authors).}\textsuperscript{1,2}}\ \ \ \
  \textbf{David Gimeno-Gomez\samethanks\textsuperscript{2}}\ \ \ \
  \textbf{Luca Ragazzi\samethanks\textsuperscript{1}}\ \ \ \ \\
  \textbf{Gianluca Moro\samethanks\textsuperscript{1}}\ \ \ \
  \textbf{Paolo Rosso\samethanks\textsuperscript{2,3}}
\\
  \textsuperscript{1}Department of Computer Science and Engineering, University of Bologna, Italy\\
  \textsuperscript{2}PRHLT Research Center, Universtitat Politècnica de València Spain,\\
  \textsuperscript{3}ValgrAI - Valencian Graduate School and Research Network of Artificial Intelligence, Spain
\\
 \normalsize{\texttt{\{paolo.italiani, l.ragazzi, gianluca.moro\}@unibo.it}, \texttt{\{dagigo1, prosso\}@dsic.upv.es}}
}

\begin{document}
\maketitle
\begin{abstract}
Women are twice as likely as men to face online harassment due to their gender. 
Despite recent advances in multimodal content moderation, most approaches still overlook the social dynamics behind this phenomenon, where perpetrators reinforce prejudices and group identity within like-minded communities.
Graph-based methods offer a promising way to capture such interactions, yet existing solutions remain limited by heuristic graph construction, shallow modality fusion, and instance-level reasoning.
In this work, we present \textsc{MemeWeaver}, an end-to-end trainable multimodal framework for detecting sexism and misogyny through a novel inter-meme graph reasoning mechanism.
We systematically evaluate multiple visual--textual fusion strategies and show that our approach consistently outperforms state-of-the-art baselines on the MAMI and EXIST benchmarks, while achieving faster training convergence.
Further analyses reveal that the learned graph structure captures semantically meaningful patterns, offering valuable insights into the relational nature of online hate.\footnote{The code is publicly available at \url{https://github.com/disi-unibo-nlp/meme-weaver}}
\end{abstract}

\section{Introduction}

Hate speech on social media is rarely an isolated act: it spreads through interactions among users.
Psychological research~\citep{walther2022online} shows that perpetrators often gain social approval by sharing content with like-minded individuals, reinforcing prejudices and group identity.
These dynamics, compounded with user anonymity~\citep{kowalski2015cyberbullying}, amplify harmful narratives and foster in-group codes and jokes, making content subtler, context-dependent, and harder to detect.

Within the spectrum of hate speech~\citep{herz2012content}, women remain one of the most disproportionately targeted groups.
They are twice as likely as men to be harassed online due to their gender~\citep{duggan2017online}, often within communities where such hateful discourses are normalized and perpetuated~\citep{fontanella2024misogyny}.
Such abuse manifests primarily as \textit{sexism}---prejudice or discrimination often rooted in stereotypes and socially accepted forms of bias~\citep{glick2001ambivalent}---and \textit{misogyny}, marked by deeper hostility, contempt, and explicit aggression~\citep{richardson2018woman}.
Persistent exposure to both has been shown to harm women's self-esteem, restrict career ambitions, and reinforce traditional gender roles~\citep{bradley2015collateral}.
The rapid dissemination of such content in online communities makes effective regulation challenging~\citep{herz2012content}, motivating the development of automated detection methods~\citep{luo2025survey}.

\begin{figure}[!t]
    \centering
    \includegraphics[width=\columnwidth]{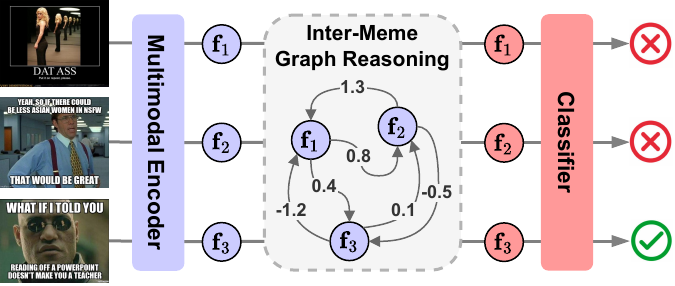}
    \caption{\textbf{Overview of \textsc{MemeWeaver}, our graph-based framework for end-to-end sexism/misogyny detection in memes.} Motivated by the social dynamics of online hate, it models batches of memes (3 shown) as inter-connected nodes for more effective learning.}
    \label{fig:memeweaver}
\end{figure}

Despite recent progress, most computational approaches overlook the social dynamics underlying online hate.
Graph-based models~\citep{zhang2019graph, rehman2025context, xu2025hyperhateprompt} provide a promising avenue by capturing inter-sample relationships.
However, existing solutions still rely on heuristic preprocessing, shallow modality fusion, and instance-level reasoning, which limit generalization and adaptability across domains.

In this paper, we introduce \textsc{MemeWeaver} (Figure~\ref{fig:memeweaver}), an end-to-end trainable framework for detecting sexism and misogyny in memes.
Our approach integrates three key ideas.
\textit{First}, memes are inherently multimodal and context-dependent: they combine text embedded in images with visual scenes, and their meaning often hinges on humor, irony, or cultural references that may obscure harmful intent~\citep{hodson2010joke, drucker2014sarcasm, plaza2024overview2, cocchieri-etal-2025-call}.
To address this, we verbalize the text content of each meme and optionally enrich it with captions generated by a multimodal LLM, capturing both surface descriptions and higher-level interpretations.
\textit{Second,} we encode text and images with CLIP-based encoders and explore alternative fusion strategies that move beyond simple concatenation, enabling richer multimodal representations.
\textit{Third}, and most importantly, we introduce a novel Inter-Meme Graph Reasoning (IMGR) mechanism that models relationships across memes within each training batch.
This design allows the system to automatically ``weave'' latent affinities between memes, capturing relational patterns without relying on heuristic graph construction.

We conduct extensive experiments on two complementary meme benchmarks: MAMI, a dataset targeting misogyny in English memes, and EXIST, a multilingual corpus focused on sexism with both English and Spanish memes.
We compare \textsc{MemeWeaver} against established baselines as well as recent large multimodal language models (LLMs).
Our main contributions are threefold:

\begin{itemize}[itemsep=1pt, topsep=1pt]
    \item \textbf{State-of-the-art results.} \textsc{MemeWeaver} consistently outperforms strong baselines on both datasets, improving over CLIP fine-tuning by $\approx$5 points on average while converging faster during training.
    \item \textbf{Fusion analysis.} We evaluate alternative text-image fusion mechanisms, showing that their effectiveness depends on dataset characteristics and that higher model complexity does not necessarily yield better performance.
    \item \textbf{Graph insights.} We analyze the learned embeddings and inter-meme graph structure, revealing semantically meaningful affinity patterns that offer insights into the relational dynamics of online hate.
\end{itemize}

\section{Related Work}

Long-standing research has addressed the detection of sexism and misogyny in online content.
This section outlines the shift from early text-only methods to recent multimodal approaches, highlighting open challenges in graph-based reasoning.

\paragraph{Sexism and Misogyny Detection}
Early work relied on statistical text features (e.g., n-grams, TF-IDF) with classical machine learning classifiers~\citep{jha2017does, anzovino2018automatic}.
Deep learning, and in particular pre-trained transformers such as BERT, improved contextual understanding~\citep{parikh2021categorizing, de2023ai}.
More recently, prompt-based techniques with LLMs have been explored for nuanced comprehension~\citep{tian2023efficient, samani2025large}.
Given the multimodal nature of social media posts~\citep{luo2025survey}, research has shifted to text-vision integration~\citep{gomez2020exploring}, especially in memes~\citep{fersini2022mami, plaza2025exist}.
Captioning-based methods~\citep{rizzi2023recognizing, plaza2024overview2}, cross-modal contrastive learning~\citep{rizzi2024pink}, multimodal LLMs~\citep{cao2022prompting, xu2025hyperhateprompt}, and prompt-based visual QA approaches such as Pro-Cap~\citep{cao2023procap} have been widely adopted.
More recent studies extend to in-the-wild videos, incorporating voice and visual cues~\citep{arcos2024sexism,plaza2025exist}.
Despite these advances, systems still struggle with implicit, context-dependent sexism and with biases in data and models, pointing to the need for richer reasoning mechanisms such as graph-based approaches.

\begin{figure*}[!t]
\centering
\includegraphics[width=\textwidth]{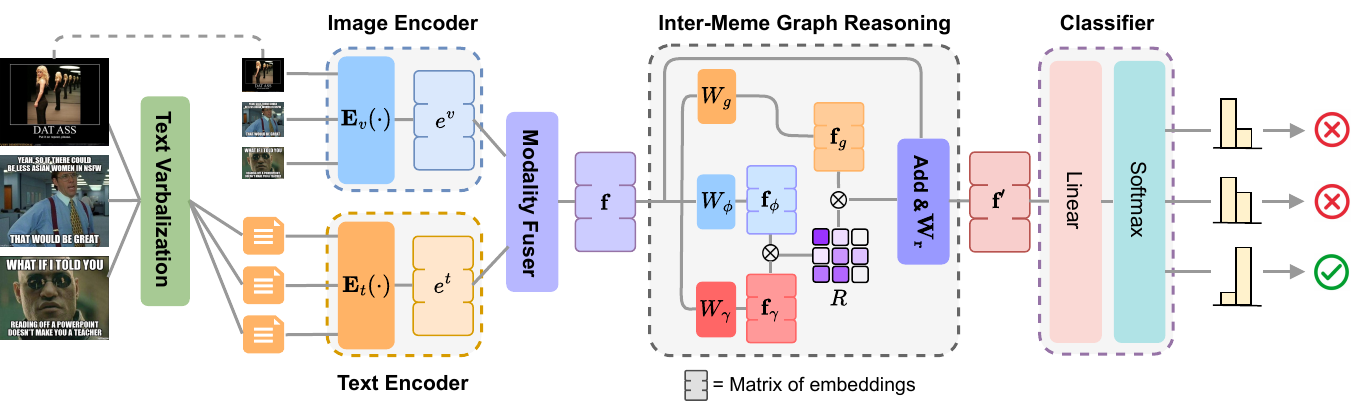}
\caption{\textbf{Architecture of \textsc{MemeWeaver}.} Each meme undergoes text extraction (OCR, optionally enriched with LLM-generated captions) and is encoded by separate text and image encoders. The embeddings are fused, refined via inter-meme graph reasoning within each batch, and classified. Example shown with batch size $m=3$.}
\label{fig:method}
\end{figure*}

\paragraph{Multimodal Graph Modeling}

Graph-based models, particularly GCNs~\citep{zhang2019graph}, are increasingly used to capture structural and semantic relations across modalities, with applications in tasks such as summarization and QA~\citep{DBLP:conf/ecai/MoroRV23, DBLP:conf/iclr/MoroRVVF24, ragazzietalTASLP25}, emotion recognition~\citep{zhu2022multimodal} and scene-text retrieval~\citep{mafla2021multi, li2024text}.
In online hate speech detection, they have also shown promise.
\citet{xu2025hyperhateprompt} combined LLMs with hypergraph learning to integrate text and visual features and capture implicit semantics, while \citet{rehman2025context} enriched unimodal representations through gated cross-attention rather than simple concatenation.
However, both approaches relied on similarity-based heuristics to define the graph.
Only \citet{hebert2024multi} introduced learnable components, though their attention-based method still depends on heuristics such as the number of user mentions in social media posts.

\paragraph{Research Gaps}
Despite recent progress, two challenges remain central.
First, most multimodal hate speech detection methods---including those targeting sexism and misogyny---rely on shallow fusion strategies, often limited to feature concatenation.
Second, existing graph-based approaches typically construct structures through heuristic rules, restricting adaptability and limiting their ability to capture inter-sample relations.
These limitations call for fully learnable frameworks that can jointly model multimodal content and social interactions in an end-to-end manner.
Our work explores whether such a design can uncover generalizable relational patterns or whether optimal configurations are inherently domain-specific.

\section{\textsc{MemeWeaver}}

We introduce \textsc{MemeWeaver}, a framework for detecting sexist and misogynistic memes, which ``weaves'' together textual and visual features while modeling relations among memes at the batch level.

\subsection{Problem Statement}
Let $\mathcal{X} = \{x_1, \dots, x_n\}$ be a multimodal dataset of memes, processed in mini-batches of size $m$.
Each instance $x_j = (t_j, v_j)$ consists of a textual component $t_j$ (OCR-extracted text, optionally enriched with an LLM-generated caption) and an image $v_j$, annotated with a binary label $y_j \in \{0,1\}$, where $1$ denotes a sexist/misogynistic meme and $0$ otherwise.
The goal is to train a multimodal classifier $f$ that, given a novel meme $(t,v)$, predicts its label $y$.

\subsection{Model Architecture}

Figure~\ref{fig:method} shows the architecture of \textsc{MemeWeaver}.
Each meme is an image that includes both a visual scene and overlaid text.
Our pipeline begins with a verbalization step, which extracts the text embedded in the image.\footnote{In the datasets used for our experiments, this text annotation is already provided by the dataset creators.}
This content may be optionally enriched with an LLM-generated caption.
The extracted text is then encoded by a text encoder, while the complete meme image is processed by a separate image encoder.
The two encoders yield modality-specific representations, which are subsequently fused into a multimodal embedding.
To capture inter-sample relationships, we weave connections among memes within each training batch, modeling the batch as a fully connected graph.
Finally, the context-enriched embeddings are fed into a classifier, with all modules trained end-to-end.

\paragraph{LLM-based Captioning}
Inspired by~\citet{xu2025hyperhateprompt}, we leverage a multimodal LLM to generate not only surface-level descriptions but also high-level interpretations of memes.
Specifically, we used Qwen2.5-VL-7B-Instruct\footnote{Run with default precision and decoding settings from the official configuration file.} and designed two prompt strategies (see Figure~\ref{fig:llm_prompts}).
The generated captions are concatenated with the OCR-extracted meme text, separated by the special token \texttt{[CPT]}; we evaluate both OCR-only and OCR+caption variants.
Examples are provided in Appendix~\ref{app:examples}.

\begin{figure}[!t]
  \begin{tcolorbox}[
      colback=orange!80!black!10!white,
      colframe=orange!80!black,
      title=Prompt A (Surface-level description),
      fonttitle=\bfseries,
      rounded corners,        
      boxrule=1pt,
      left=6pt, right=6pt, top=4pt, bottom=4pt,
      width=\columnwidth,
      fontupper=\ttfamily\footnotesize\itshape,  
      enhanced jigsaw         
    ]
    Describe this image without including what text reads and credit sources.
  \end{tcolorbox}

  \begin{tcolorbox}[
      colback=customblue2!10!white,
      colframe=customblue2,
      title=Prompt B (Multimodal semantic inference),
      fonttitle=\bfseries,
      rounded corners,        
      boxrule=1pt,
      left=6pt, right=6pt, top=4pt, bottom=4pt,
      width=\columnwidth,
      fontupper=\ttfamily\footnotesize\itshape,  
      enhanced jigsaw         
    ]
    You are a helpful assistant designed to detect \{HATE\_TYPE\} expressions or behaviours in a meme, i.e., it is \{HATE\_TYPE\} itself, describes a \{HATE\_TYPE\} situation or criticizes a \{HATE\_TYPE\} behaviour. Infer the implicit semantic information of the meme, considering that it may or may not contain \{HATE\_TYPE\} content. Please be concise (no more than three sentences) while including all relevant information.
  \end{tcolorbox}
  \caption{\textbf{Prompts used for LLM-based meme captioning.} The \texttt{HATE\_TYPE} placeholder was set to \textit{misogynistic} or \textit{sexist} depending on the downstream dataset.}
  \label{fig:llm_prompts}
\end{figure}

\paragraph{Text Encoder}
Given a batch of memes, we process their textual components \(\{t_{1}, \dots, t_{m}\}\) with a transformer-based text encoder \(\mathbf{E}_{t}(\cdot)\), yielding hidden representations:
\begin{equation}
\begin{split}
    e_j^t = \mathbf{E}_t(t_j), \quad j = 1, \dots, m \\
    e^t = [e_1^t, \ldots, e_m^t] \in \mathbb{R}^{m \times \tfrac{d}{2}}
\end{split}
\end{equation}
where \(d=1536\) denotes the overall multimodal hidden dimensionality, so that each encoder outputs vectors of size \(d/2\).

\paragraph{Image Encoder}
Similarly, given the same batch of memes, we process the meme images \(\{v_{1}, \dots, v_{m}\}\) with an image encoder \(\mathbf{E}_{v}(\cdot)\), yielding hidden representations:
\begin{equation}
\begin{split}
    e_j^v &= \mathbf{E}_v(v_j), \quad j = 1, \dots, m \\
    e^v &= [e_1^v, \ldots, e_m^v] \in \mathbb{R}^{m \times \tfrac{d}{2}}
\end{split}
\end{equation}
with the same dimensionality convention as in the text encoder.
Both the text and image encoders are initialized from CLIP and fine-tuned jointly with the rest of the architecture.

\paragraph{Modality Fuser}
In addition to simple concatenation, we consider two fusion mechanisms to combine the text and image embeddings into unified multimodal representations 
$\mathbf{f} = [f_1, \ldots, f_m]$.

\textit{(1) Multi-modal Factorized Bilinear (MFB) pooling}~\citep{DBLP:conf/iclr/KimOLKHZ17} 
projects each modality into a shared space and applies element-wise interaction:
\begin{equation}
    f_j = ( e_j^tU^\top) \circ (e_j^vV^\top ), \quad j=1,\dots,m
\end{equation}
where $U,V \in \mathbb{R}^{\tfrac{d}{2} \times d}$ are trainable projection matrices and $\circ$ denotes the Hadamard product.

\textit{(2) Gated Multimodal Unit (GMU)}~\citep{DBLP:conf/iclr/OvalleSMG17} learns a gating vector to adaptively control each modality's contribution:
\begin{equation}
\begin{split}
  z_j &= \sigma\!\left([e_j^t, e_j^v]U_z \right), \\
  f_j &= z_j \odot \tanh( e_j^vU_v^\top) \\ &\quad
       + (1 - z_j) \odot \tanh(e_j^tU_t^\top), \quad j=1,\dots,m
\end{split}
\end{equation}
where $U_t, U_v \in \mathbb{R}^{\tfrac{d}{2} \times d}$ and $U_z \in \mathbb{R}^{d \times d}$ are trainable matrices, 
$\sigma$ is the sigmoid, $[\cdot,\cdot]$ denotes concatenation, and $\odot$ denotes multiplication.

\paragraph{Inter-Meme Graph Reasoning (IMGR)}
Inspired by prior work on relational reasoning~\citep{li2019visual,mafla2021multi}, we enhance multimodal representations by modeling relationships \textit{across} memes in a batch, rather than within a single instance (e.g., linking image regions or textual tokens).
We compute an affinity matrix $R \in \mathbb{R}^{m \times m}$ capturing pairwise meme similarities:
\begin{equation}
    R = (\mathbf{f} W_\phi)(\mathbf{f} W_\gamma)^\top
\end{equation}
where $\mathbf{f} = [f_1, \ldots, f_m] \in \mathbb{R}^{m \times d}$ are fused embeddings and $W_\phi, W_\gamma \in \mathbb{R}^{d \times d}$ are trainable projections.
This yields a fully connected graph $G = (\mathbf{f}, R)$, with nodes $f_i$ (memes) and edge weights $R_{ij}$ (their relations).
Following~\citet{li2019visual}, we then apply a graph-based message passing layer with residual connections:
\begin{equation}
    \mathbf{f}' = (\hat{R} \mathbf{f} W_g)W_r + \mathbf{f}
\end{equation}
where $W_r, W_g \in \mathbb{R}^{d \times d}$ are trainable projections and $\hat{R}\! =\! \frac{R}{m}$ is the uniform rescaled affinity matrix.
The final representations $\mathbf{f}' = [f'_1, \ldots, f'_m]$ are enriched with batch-level context, enabling each meme to incorporate information from its peers.

\paragraph{Classifier and Loss Function}
The classifier consists of a linear projection $W_c \in \mathbb{R}^{d \times 2}$ producing logits, followed by a softmax that yields probabilities $P = [p_1, \ldots, p_m] \in \mathbb{R}^{m \times 2}$, where each $p_j \in \mathbb{R}^2$ is the predicted distribution for meme $j$.
All modules of our architecture are trained end-to-end using cross-entropy loss:
\begin{equation}
\mathcal{L}_{ce} = -\frac{1}{m} \sum_{j=1}^m 
    \sum_{c \in \{0,1\}} \mathbf{1}[y_j = c] \, \log p_{j,c}
\end{equation}
where $y_j \in \{0,1\}$ is the gold label for instance $j$.

\section{Experimental Setup}

\subsection{Datasets}

To capture domain and cultural diversity, we evaluate our method on two distinct multimodal meme datasets for misogyny and sexism detection, framing the task as binary classification following prior approaches~\citep{cao2023procap, xu2025hyperhateprompt}.

\textbf{(1) MAMI}~\citep{fersini2022mami} is a curated SemEval dataset focused specifically on \textit{misogyny} in memes, collected from Twitter, Reddit, and meme-focused websites.
It provides fine-grained subtype annotations such as shaming, stereotyping, objectification, and incitement to violence; we assign the misogyny label if a meme belongs to any subtype.
This widely used benchmark serves as our primary dataset for misogyny detection in English memes.

\textbf{(2) EXIST}~\citep{plaza2025exist} is a CLEF shared task on \textit{sexism} detection across social media, including tweets and videos.
Here we focus on Task 2, targeting memes.
The dataset was built from 250 sexism-related queries on Google Images, and includes subtype labels such as ideological inequality, stereotyping and dominance, objectification, and sexual violence.
Since each meme was annotated by three annotators, we derived the final sexism label via majority voting.
Unlike MAMI, EXIST contains memes in both English and Spanish, allowing cross-lingual and cross-cultural evaluation.

\begin{table}[!t]
\centering
\begin{adjustbox}{max width=\columnwidth}
\begin{tabular}{lrrrrr}
 \toprule
 \multirow{2}{*}[-3pt]{\textbf{Dataset}} & & \multicolumn{3}{c}{\textbf{No. of Samples \textcolor{red!50!black}{(\% Misogyny/Sexism)}}} \\ \cmidrule{3-5}
 & &  \multicolumn{1}{c}{\textit{Training}} & \multicolumn{1}{c}{\textit{Validation}} & \multicolumn{1}{c}{\textit{Test}}\\ \midrule
 MAMI & & 9,000 {\small\textcolor{red!50!black}{(50.0\%)}} & 1,000 {\small\textcolor{red!50!black}{(50.0\%)}} & 1,000 {\small\textcolor{red!50!black}{(50.0\%)}} \\
 EXIST & & 3,235 {\small\textcolor{red!50!black}{(65.7\%)}} & 404 {\small\textcolor{red!50!black}{(67.3\%)}} & 405 {\small\textcolor{red!50!black}{(65.7\%)}} \\
 \bottomrule
\end{tabular}
\end{adjustbox}
\caption{\textbf{Dataset statistics.} We highlight the percentage of sexist and misogynistic memes in each split.}
\label{tab:datasets}
\end{table}

\begin{figure}[!t]
    \centering
    \includegraphics[width=\columnwidth]{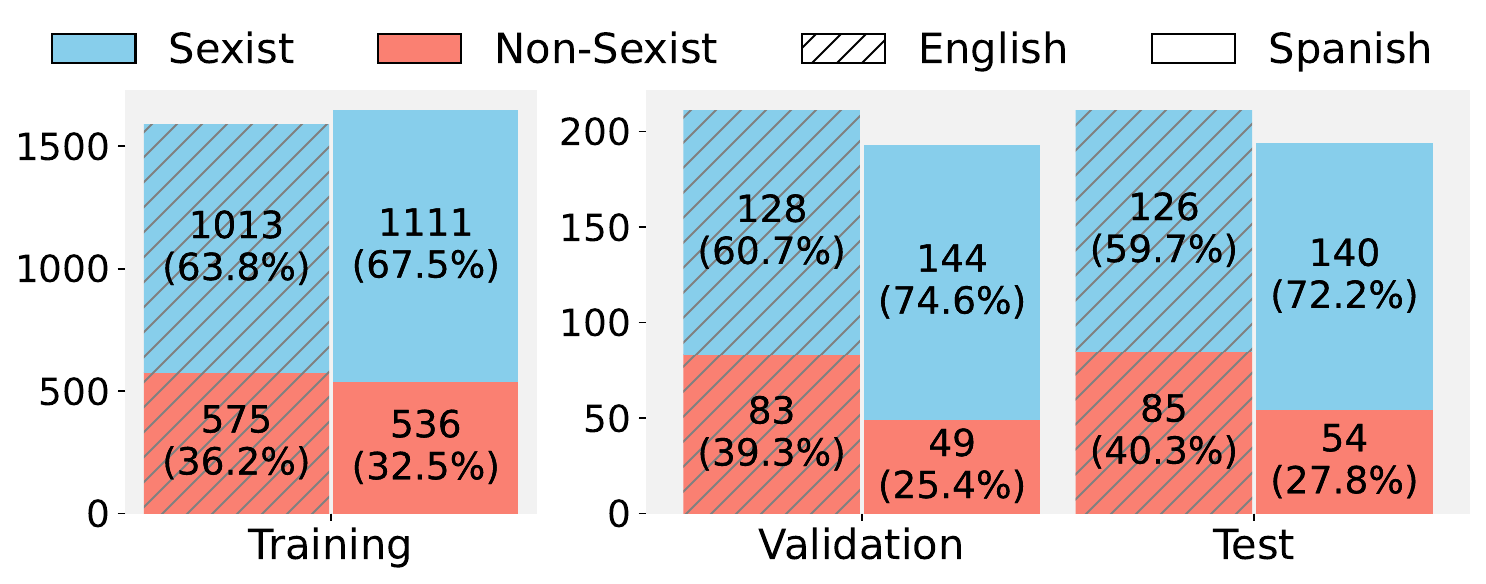}
    \caption{\textbf{Language distribution in EXIST.} Ratio of sexist vs. non-sexist memes across splits.}
    \label{fig:exist}
\end{figure}

Table~\ref{tab:datasets} shows the dataset distributions.
MAMI is balanced (50\% misogynistic memes), while EXIST is skewed toward sexist content ($\approx$66\%).
Figure~\ref{fig:exist} further breaks down EXIST by language and class across splits: the training set is relatively balanced, but the skew becomes more pronounced in the validation and test sets, especially for Spanish.
For MAMI we rely on the official splits, whereas for EXIST we define a custom 80/10/10 train/val/test split, since no validation set is provided and the official test set is not publicly available.\footnote{This ensures consistent internal evaluation, though results are not directly comparable to leaderboard scores.}
Both datasets already come with OCR-extracted text.

\subsection{Baselines}

We use the widely adopted multimodal model CLIP~\citep{radford2021learning}\footnote{HuggingFace checkpoint: openai/clip-vit-large-patch14} as the main backbone of \textsc{MemeWeaver}.
We choose CLIP over larger multimodal LLMs because its lightweight design allows for larger effective batch sizes, thereby facilitating richer inter-meme relationships within IMGR.
Unlike for EXIST, we use the original test set for MAMI, and thus report results against established baselines: EF-CaTrBERT~\citep{khan2021exploiting}, PromptHate~\citep{cao2022prompting}, Pro-Cap~\citep{cao2023procap}, and HyperHatePrompt~\citep{xu2025hyperhateprompt}.
In addition, we compare against recent multimodal LLMs, such as Qwen2.5-VL~\citep{bai2025qwen2}, Phi-4~\citep{abdin2024phi}, and GPT-4o-mini~\citep{hurst2024gpt}, evaluated in a zero-shot setting using instruction-following prompts (details are provided in Appendix~\ref{app:prompts}).

\begin{table}[!t]
\centering
\begin{adjustbox}{width=\linewidth}
\begin{tabular}{l|c|cc}
\toprule
\textbf{Setting} & \textbf{Range} & \textbf{MAMI} & \textbf{EXIST} \\
\midrule
Training batch & [10, 180] & 20 & 64 \\
Inference batch & [10, 180] & 41 & 27 \\
Class. threshold & [0.010, 0.999] & 0.657 & 0.434 \\
Modality fusion & N/A & MFB & Concat \\
Image captioning & N/A & – & Prompt A \\
\bottomrule
\end{tabular}
\end{adjustbox}
\caption{\textbf{Optimal settings.} Ranges denote search intervals; final values were selected on the validation sets.}
\label{table:hyperparams}
\end{table}

\begin{table*}[!t]
\centering
\small
\begin{adjustbox}{max width=\textwidth}
\begin{tabular}{l|ccc|ccc|ccc|ccc}
\toprule
& \multicolumn{3}{c|}{\textbf{MAMI}} & \multicolumn{9}{c}{\textbf{EXIST}} \\
\multicolumn{1}{c|}{} & \multicolumn{3}{c|}{\textit{English}} & \multicolumn{3}{c}{\textit{All}} & \multicolumn{3}{c}{\textit{English}} & \multicolumn{3}{c}{\textit{Spanish}} \\
\cmidrule(lr{0.5em}){2-4}
\cmidrule(lr{0.5em}){5-7}
\cmidrule(lr{0.5em}){8-10}
\cmidrule(lr{0.5em}){11-13}
\textbf{Method} & \textbf{Acc} & \textbf{F1} & \textbf{AUC} & \textbf{Acc} & \textbf{F1} & \textbf{AUC} & \textbf{Acc} & \textbf{F1} & \textbf{AUC} & \textbf{Acc} & \textbf{F1} & \textbf{AUC} \\
\midrule
Phi-4 & 69.8 & 70.0 & - & 61.4 & 62.5 & - & 66.4 & 65.8 & - & 58.3 & 56.3 & - \\
GPT-4o-mini & 71.7 & 73.1 & - & 64.4 & 64.4 & - & 70.1 & 70.1 & - & 57.8 & 58.2 & - \\
Qwen2.5-VL & 74.4 & 74.8 & - & 71.4 & \underline{70.0} & - & 72.5 & \underline{72.1} & - & 70.1 & 66.7 & - \\
\midrule
\rowcolor{mygray}\multicolumn{13}{l}{\textsc{MemeWeaver}} \\
\hspace{15pt}w/\hspace{4.5pt} IMGR & \textbf{77.6}* & \textbf{77.4}* & \underline{83.4}* & 73.3 & 67.9* & \underline{76.9}* & 71.6 & 68.4* & \underline{77.0}* & 75.3 & 66.3* & \textbf{76.5}* \\
\hspace{15pt}w/o IMGR & 71.9 & 71.8 & 79.4 & 71.1 & 60.5 & 71.2 & 69.2 & 61.6 & 74.8 & 73.2 & 58.3 & 65.9 \\
\hline
\rowcolor{mygray}\multicolumn{13}{l}{$+$ Prompt A} \\
\hspace{15pt}w/\hspace{4.5pt} IMGR & 74.8 & 74.4 & 80.1 & \textbf{76.3} & \textbf{71.6}* & 75.7* & \textbf{76.3}* & \textbf{73.6}* & 72.9 & \underline{75.3} & \underline{67.2} & 66.4 \\
\hspace{15pt}w/o IMGR & 69.2 & 68.9 & 77.5 & 72.1 & 61.8 & 73.5 & 68.2 & 59.5 & 75.1 & \textbf{76.3} & 64.4 & 72.1 \\
\hline
\rowcolor{mygray}\multicolumn{13}{l}{$+$ Prompt B} \\
\hspace{15pt}w/\hspace{4.5pt} IMGR & \underline{76.3}* & \underline{76.6}* & \textbf{83.5}* & \underline{73.6} & 67.3 & \textbf{78.9}* & \underline{73.9} & 71.3 & \textbf{80.4}* & 73.2 & 59.1 & \underline{76.4} \\
\hspace{15pt}w/o IMGR & 72.7 & 71.9 & 78.8 & 73.3 & 69.7 & 74.3 & 73.5 & 71.0 & 76.3 & 73.2 & \textbf{67.7} & 72.3 \\
\bottomrule
\end{tabular}
\end{adjustbox}
\caption{\textbf{Comparison to state of the art on test sets.} Best results in bold, second best underlined. * denotes statistically significant improvements ($\alpha = 0.05$,~\citealp{koehn2004statistical}). Ablation studies confirm the importance of IMGR.}
\label{tab:main_results}
\end{table*}

\begin{table}[!t]
\centering
\small
\begin{adjustbox}{width=\columnwidth}
\begin{tabular}{lccc}
\toprule
& \textbf{Acc} & \textbf{F1} & \textbf{AUC} \\
\midrule
\multicolumn{4}{l}{\textbf{Prior Work}} \\
\hspace{10pt}EF-CaTrBERT~\citeyearpar{khan2021exploiting} & 67.8 & 67.2 & 74.9 \\
\hspace{10pt}PromptHate~\citeyearpar{cao2022prompting} & 71.1 & 70.8 & 80.8 \\
\hspace{10pt}Pro-Cap~\citeyearpar{cao2023procap} & 73.3 & 72.4 & 83.2 \\
\hspace{10pt}HyperHatePrompt~\citeyearpar{xu2025hyperhateprompt} & \underline{75.3} & \underline{75.1} & \textbf{84.3} \\
\midrule
\textsc{MemeWeaver} (Ours) & \textbf{77.6} & \textbf{77.4} & \underline{83.4} \\
\bottomrule
\end{tabular}
\end{adjustbox}
\caption{\textbf{Our best \textsc{MemeWeaver} vs. state of the art on MAMI.} Best scores in bold, second-best underlined.}
\label{tab:sota}
\end{table}

\subsection{Implementation Details}

\paragraph{Environment}
All experiments run on a workstation equipped with an NVIDIA GeForce RTX 3090 GPU (24~GB VRAM), 64~GB system RAM, and an Intel® Core™ i9-10900X CPU @ 3.70GHz.
Our implementation builds on Python 3.10.12, PyTorch 2.2.1, and HuggingFace Transformers 4.44.0.

\paragraph{Multilingual Setup}
For EXIST, we translated the Spanish split into English using Google Translate, since CLIP is English-only.
Given the short and simple nature of the texts, automatic translation is highly reliable.
We manually inspected the entire translated set and found no systematic errors or meaning distortions, ensuring negligible translation bias.
Additionally, we conducted an automatic evaluation of the translation quality (see details in Appendix~\ref{app:translation_quality}).
This choice enables a unified backbone across datasets, avoiding confounding effects from heterogeneous multilingual encoders and allowing a fair assessment of IMGR.
It also represents a practical strategy for handling multilingual memes when only monolingual encoders are available.

\paragraph{Method Settings}
For the EXIST dataset, which is relatively low-resource in terms of available training data---thus also simulating a more realistic scenario with limited supervision~\citep{Moro_Ragazzi_2022, MORO2023126356}---we initialize from the checkpoint pretrained on MAMI before fine-tuning.
All models are trained for 3 epochs with the AdamW optimizer (learning rate 5e-6) and a linear scheduler, using a fixed random seed of 42 for reproducibility.
Training and inference batch sizes are tuned on the validation set, as they directly determine how many instances can interact within the IMGR module and thus critically impact \textsc{MemeWeaver}'s performance.
Finally, rather than adopting the default $0.5$ classification threshold, we follow \citet{xu2025hyperhateprompt} and select the cutoff that maximizes F1 on the validation set.
Table~\ref{table:hyperparams} summarizes all final hyperparameters, including not only optimization settings but also the selected fusion strategy and the use of LLM-based captioning along with the best-performing prompt.

\paragraph{Evaluation Metrics}
For each model, we retain the checkpoint with the highest macro-F1 score on the validation set.
We report three complementary metrics: Accuracy (Acc), macro-F1, and Area Under the Curve (AUC).
Among these, macro-F1 serves as our primary metric, as it equally weights both classes and is therefore more reliable under the label imbalance in the EXIST dataset.\footnote{For generative LLMs, which output hard labels rather than probability distributions, AUC is not reported.}

\begin{figure*}[!t]
  \centering
    \begin{minipage}{0.8\linewidth}
      \centering
      \footnotesize
      \colorbox{customblue!20}{%
        \tikz[baseline=-0.5ex]\draw[fill=customblue,draw=customblue] (0,0) circle (3pt);~Non-Sexist/Misogynistic%
      }
      \quad
      \colorbox{customred!20}{%
        \tikz[baseline=-0.5ex]\draw[fill=customred,draw=customred] (0,0) circle (3pt);~Sexist/Misogynistic%
      }
  \end{minipage}
  \begin{subfigure}[b]{0.24\linewidth}
    \centering
    \includegraphics[width=\linewidth]{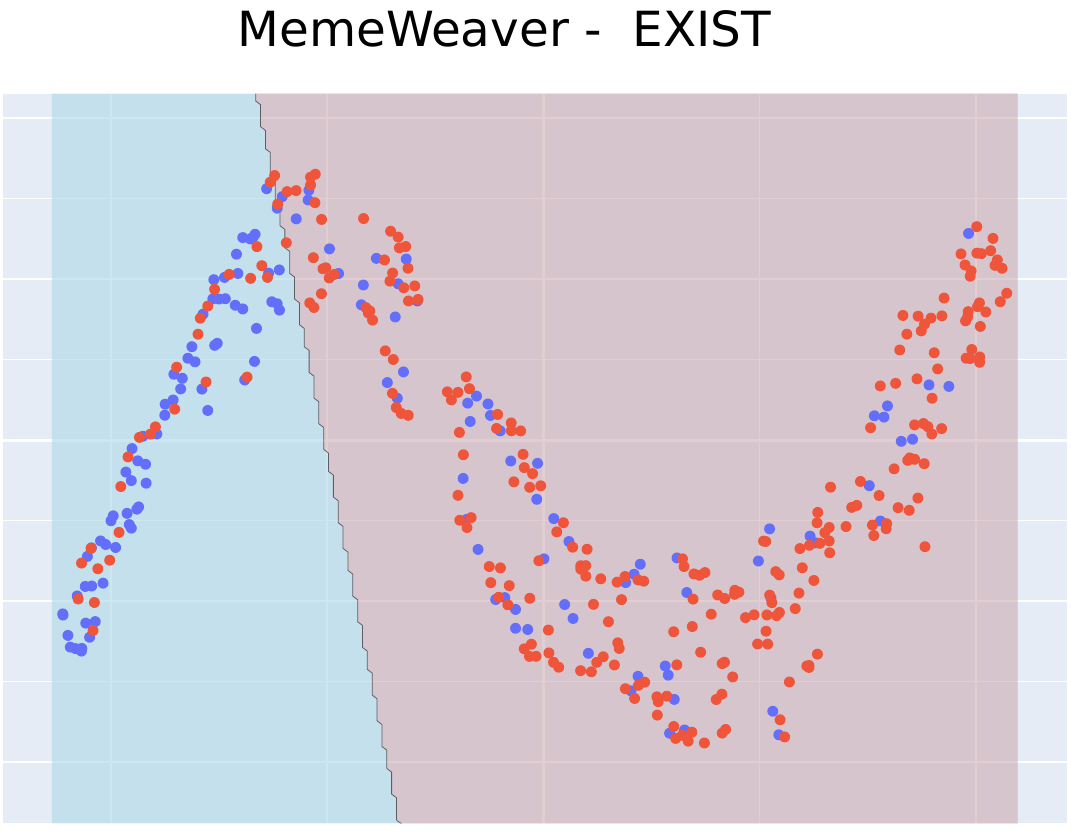}
    \caption*{\footnotesize \textbf{Acc} = 74.7}
    \label{fig:exist-graph}
  \end{subfigure}\hfill
  \begin{subfigure}[b]{0.24\linewidth}
    \centering
    \includegraphics[width=\linewidth]{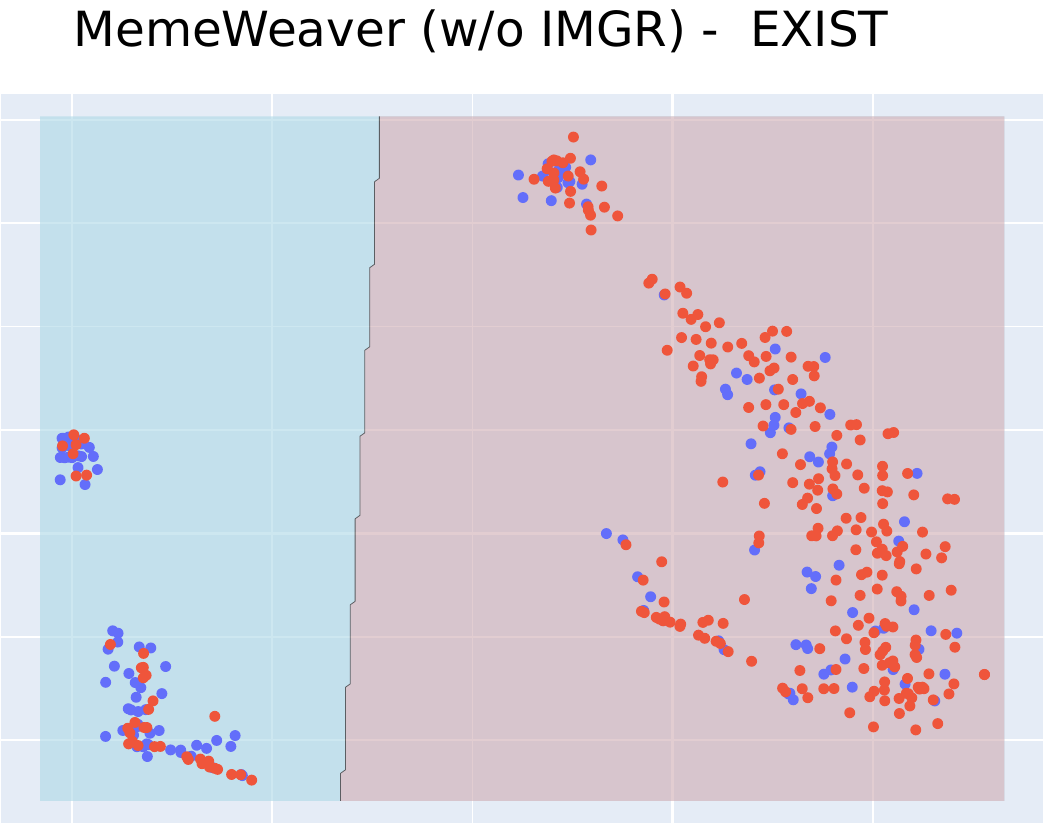}
    \caption*{\footnotesize \textbf{Acc} = 72.1}
    \label{fig:exist-nograph}
  \end{subfigure}\hfill
  \begin{subfigure}[b]{0.24\linewidth}
    \centering
    \includegraphics[width=\linewidth]{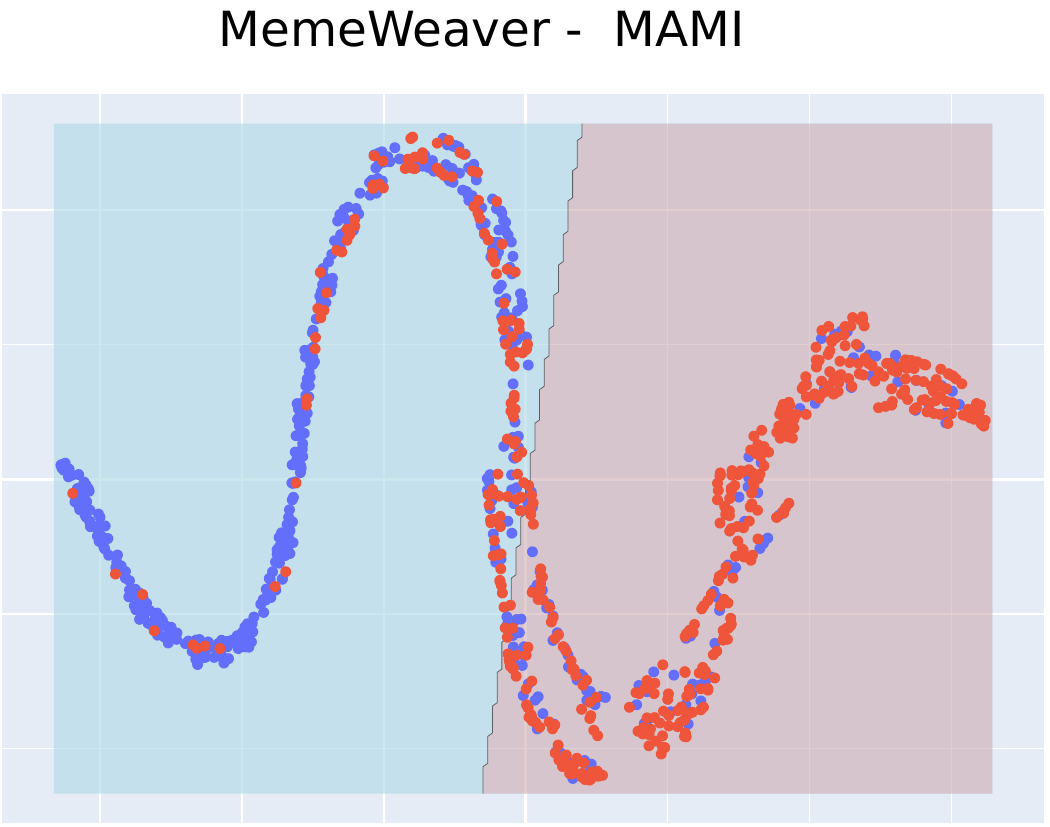}
    \caption*{\footnotesize \textbf{Acc} = 75.2}
    \label{fig:mami-graph}
  \end{subfigure}\hfill
  \begin{subfigure}[b]{0.24\linewidth}
    \centering
    \includegraphics[width=\linewidth]{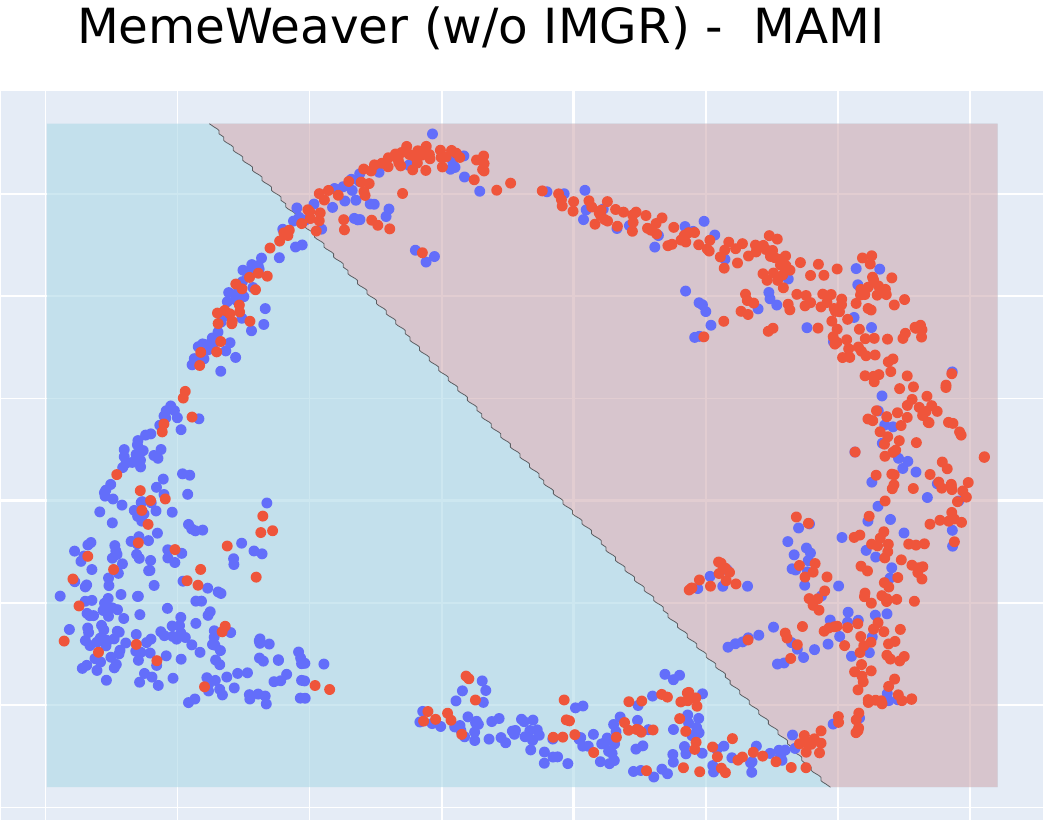}
    \caption*{\footnotesize \textbf{Acc} = 74.7}
    \label{fig:mami-nograph}
  \end{subfigure}
  \caption{\textbf{Embedding space analysis.} 2D visualizations of MAMI and EXIST test set embeddings $\mathbf{f}'$ and $\mathbf{f}$, generated with and without \textsc{MemeWeaver}'s IMGR, respectively. Decision boundaries from a linear classifier show higher accuracy on $\mathbf{f}'$ (see scores below plots), underscoring the benefit of IMGR's representations.}
  \label{fig:emb_comparison}
\end{figure*}

\section{Results \& Discussion}

\paragraph{Performance Comparison}

Table~\ref{tab:main_results} shows that \textsc{MemeWeaver} with IMGR consistently outperforms ablated variants without graph reasoning, particularly on MAMI and EXIST-English, with gains of 7-14 F1 points depending on the configuration, yielding more discriminative feature representations across memes.
This effectiveness is further supported by Table~\ref{tab:sota}, where we establish a new state of the art on MAMI over prior work.

Performance differences across datasets reflect the subtler, more complex nature of sexist discourse in EXIST compared to the more explicit, hate-driven memes in MAMI.
This may also explain why incorporating complementary image captioning is more beneficial in the sexism case.
Yet, the limited impact of semantic captioning (Prompt B) aligns with the weaker performance of our zero-shot LLM experiments, suggesting that even these powerful systems struggle with context-dependent social phenomena.
This contrasts with prior studies that emphasized captioning as crucial for multimodal hate detection~\cite{cao2022prompting,cao2023procap}. Furthermore, beyond improved overall performance, our method exhibits greater robustness across languages compared to LLMs. Specifically, LLMs show a substantial cross-lingual discrepancy, with an average F1 gap of 8.9{\small $\pm$2.7} between the English and Spanish EXIST partitions.

Overall, the results highlight the central role of learnable graph reasoning and robust multimodal fusion, which proved to adapt more effectively than both LLMs and hand-crafted heuristics.

\begin{figure}[!t]
    \centering
    
    \begin{tikzpicture}
    \begin{axis}[
      hide axis,
      xmin=0, xmax=1, ymin=0, ymax=1,
      legend columns=2,
      legend cell align=left,
      legend style={
        at={(0.5,1.05)},
        anchor=south,
        draw=none,
        fill=none,
        /tikz/column sep=0.4em, 
        font=\footnotesize,
      },
      legend image post style={xshift=-0.25em}, 
    ]
    \addlegendimage{mamiColor, line width=3pt}
    \addlegendentry{MAMI}
    \addlegendimage{black, solid, line width=3pt}
    \addlegendentry{\textsc{MemeWeaver}}
    
    \addlegendimage{existColor, line width=3pt}
    \addlegendentry{EXIST}
    \addlegendimage{black, dotted, line width=3pt}
    \addlegendentry{\textsc{MemeWeaver} (w/o IMGR)}
    \end{axis}
    \end{tikzpicture}
    \begin{subfigure}[H]{\columnwidth}
        \begin{tikzpicture}
            \begin{axis}[
                width=\textwidth,
                height=4.5cm,
                ymajorgrids=true,
                grid=both,
                grid style=dashed,
                axis background/.style={fill=blue!4},
                xmin=0, xmax=1,
                ymin=0.001, ymax=0.762,
                xlabel={Normalized Global Step},
                ylabel={Training Loss},
                xlabel style={font=\footnotesize},
    ylabel style={font=\footnotesize},
                every tick label/.append style={font=\fontsize{8}{8}\selectfont},
            ]
            \addplot[color=existColor, line width=1.5pt, dotted] coordinates {
                (0.010,0.6497)
                (0.020,0.6152)
                (0.029,0.6297)
                (0.039,0.5692)
                (0.049,0.5797)
                (0.059,0.6277)
                (0.069,0.6238)
                (0.078,0.6283)
                (0.088,0.5812)
                (0.098,0.5549)
                (0.108,0.5865)
                (0.118,0.5887)
                (0.127,0.5789)
                (0.137,0.5797)
                (0.147,0.5498)
                (0.157,0.5360)
                (0.167,0.5568)
                (0.176,0.5414)
                (0.186,0.6070)
                (0.196,0.5493)
                (0.206,0.5338)
                (0.216,0.5400)
                (0.225,0.5100)
                (0.235,0.5228)
                (0.245,0.5519)
                (0.255,0.4806)
                (0.265,0.5455)
                (0.275,0.5209)
                (0.284,0.5072)
                (0.294,0.4897)
                (0.304,0.5079)
                (0.314,0.4831)
                (0.324,0.4651)
                (0.333,0.4442)
                (0.343,0.4549)
                (0.353,0.4989)
                (0.363,0.4914)
                (0.373,0.4643)
                (0.382,0.4682)
                (0.392,0.4556)
                (0.402,0.4791)
                (0.412,0.4236)
                (0.422,0.4545)
                (0.431,0.3942)
                (0.441,0.4053)
                (0.451,0.3982)
                (0.461,0.4293)
                (0.471,0.4336)
                (0.480,0.4443)
                (0.490,0.4317)
                (0.500,0.4426)
                (0.510,0.4194)
                (0.520,0.4078)
                (0.529,0.4025)
                (0.539,0.4019)
                (0.549,0.4162)
                (0.559,0.3735)
                (0.569,0.3739)
                (0.578,0.3749)
                (0.588,0.3820)
                (0.598,0.3802)
                (0.608,0.3870)
                (0.618,0.3761)
                (0.627,0.3666)
                (0.637,0.3426)
                (0.647,0.3656)
                (0.657,0.3343)
                (0.667,0.3908)
                (0.676,0.3670)
                (0.686,0.3852)
                (0.696,0.3433)
                (0.706,0.4013)
                (0.716,0.3664)
                (0.725,0.3784)
                (0.735,0.3411)
                (0.745,0.3435)
                (0.755,0.3211)
                (0.765,0.3427)
                (0.775,0.3335)
                (0.784,0.3574)
                (0.794,0.3481)
                (0.804,0.3592)
                (0.814,0.3524)
                (0.824,0.3538)
                (0.833,0.3114)
                (0.843,0.3388)
                (0.853,0.3248)
                (0.863,0.3429)
                (0.873,0.3385)
                (0.882,0.3392)
                (0.892,0.3596)
                (0.902,0.3485)
                (0.912,0.3044)
                (0.922,0.3296)
                (0.931,0.3428)
                (0.941,0.2864)
                (0.951,0.3360)
                (0.961,0.3153)
                (0.971,0.3006)
                (0.980,0.3051)
                (0.990,0.3075)
                (1.000,0.3353)
            };
            \addplot[color=existColor, line width=1.5pt, solid] coordinates {
                (0.010,0.7466)
                (0.020,0.5985)
                (0.029,0.5837)
                (0.039,0.5226)
                (0.049,0.5254)
                (0.059,0.5935)
                (0.069,0.5444)
                (0.078,0.6181)
                (0.088,0.5354)
                (0.098,0.5076)
                (0.108,0.5260)
                (0.118,0.4815)
                (0.127,0.4725)
                (0.137,0.4811)
                (0.147,0.4397)
                (0.157,0.4289)
                (0.167,0.4919)
                (0.176,0.4436)
                (0.186,0.4737)
                (0.196,0.4511)
                (0.206,0.3597)
                (0.216,0.3536)
                (0.225,0.2871)
                (0.235,0.3730)
                (0.245,0.3117)
                (0.255,0.2790)
                (0.265,0.3241)
                (0.275,0.3214)
                (0.284,0.3479)
                (0.294,0.2747)
                (0.304,0.2545)
                (0.314,0.2282)
                (0.324,0.1740)
                (0.333,0.1655)
                (0.343,0.1997)
                (0.353,0.2528)
                (0.363,0.1437)
                (0.373,0.2023)
                (0.382,0.1558)
                (0.392,0.1539)
                (0.402,0.2073)
                (0.412,0.1129)
                (0.422,0.1594)
                (0.431,0.1461)
                (0.441,0.1585)
                (0.451,0.0996)
                (0.461,0.1108)
                (0.471,0.1250)
                (0.480,0.1222)
                (0.490,0.1452)
                (0.500,0.1224)
                (0.510,0.1104)
                (0.520,0.0843)
                (0.529,0.1152)
                (0.539,0.1425)
                (0.549,0.1721)
                (0.559,0.0786)
                (0.569,0.0752)
                (0.578,0.0737)
                (0.588,0.1032)
                (0.598,0.0635)
                (0.608,0.1385)
                (0.618,0.0517)
                (0.627,0.0499)
                (0.637,0.0349)
                (0.647,0.0934)
                (0.657,0.0903)
                (0.667,0.0822)
                (0.676,0.0611)
                (0.686,0.0772)
                (0.696,0.0373)
                (0.706,0.0275)
                (0.716,0.0400)
                (0.725,0.0170)
                (0.735,0.0785)
                (0.745,0.0596)
                (0.755,0.0560)
                (0.765,0.0331)
                (0.775,0.0781)
                (0.784,0.0203)
                (0.794,0.0186)
                (0.804,0.0680)
                (0.814,0.0286)
                (0.824,0.0724)
                (0.833,0.0243)
                (0.843,0.0864)
                (0.853,0.0381)
                (0.863,0.0867)
                (0.873,0.1839)
                (0.882,0.0576)
                (0.892,0.0383)
                (0.902,0.0262)
                (0.912,0.0015)
                (0.922,0.0242)
                (0.931,0.0301)
                (0.941,0.0038)
                (0.951,0.0698)
                (0.961,0.0177)
                (0.971,0.0285)
                (0.980,0.0327)
                (0.990,0.0050)
                (1.000,0.0657)
            };
            \addplot[color=mamiColor, line width=1.5pt, dotted] coordinates {
                (0.010,0.6930)
                (0.020,0.6915)
                (0.030,0.6875)
                (0.040,0.6834)
                (0.050,0.6777)
                (0.060,0.6721)
                (0.070,0.6669)
                (0.080,0.6618)
                (0.090,0.6592)
                (0.100,0.6533)
                (0.110,0.6424)
                (0.120,0.6363)
                (0.130,0.6326)
                (0.140,0.6250)
                (0.150,0.6212)
                (0.160,0.6166)
                (0.170,0.6164)
                (0.180,0.6137)
                (0.190,0.6107)
                (0.200,0.6004)
                (0.210,0.5913)
                (0.220,0.5824)
                (0.230,0.5795)
                (0.240,0.5770)
                (0.250,0.5717)
                (0.260,0.5732)
                (0.270,0.5668)
                (0.280,0.5640)
                (0.290,0.5502)
                (0.300,0.5541)
                (0.310,0.5401)
                (0.320,0.5350)
                (0.330,0.5265)
                (0.340,0.5208)
                (0.350,0.5166)
                (0.360,0.5089)
                (0.370,0.5130)
                (0.380,0.4981)
                (0.390,0.5052)
                (0.400,0.5020)
                (0.410,0.4980)
                (0.420,0.4831)
                (0.430,0.4842)
                (0.440,0.4804)
                (0.450,0.4751)
                (0.460,0.4793)
                (0.470,0.4606)
                (0.480,0.4537)
                (0.490,0.4550)
                (0.500,0.4595)
                (0.510,0.4583)
                (0.520,0.4451)
                (0.530,0.4423)
                (0.540,0.4436)
                (0.550,0.4315)
                (0.560,0.4378)
                (0.570,0.4326)
                (0.580,0.4199)
                (0.590,0.4247)
                (0.600,0.4230)
                (0.610,0.4254)
                (0.620,0.4100)
                (0.630,0.4308)
                (0.640,0.4039)
                (0.650,0.4142)
                (0.660,0.4000)
                (0.670,0.4001)
                (0.680,0.4010)
                (0.690,0.4123)
                (0.700,0.3759)
                (0.710,0.3955)
                (0.720,0.3854)
                (0.730,0.3897)
                (0.740,0.3840)
                (0.750,0.3942)
                (0.760,0.3860)
                (0.770,0.3597)
                (0.780,0.3753)
                (0.790,0.3720)
                (0.800,0.3800)
                (0.810,0.3711)
                (0.820,0.3744)
                (0.830,0.3679)
                (0.840,0.3543)
                (0.850,0.3645)
                (0.860,0.3638)
                (0.870,0.3457)
                (0.880,0.3748)
                (0.890,0.3689)
                (0.900,0.3647)
                (0.910,0.3576)
                (0.920,0.3648)
                (0.930,0.3586)
                (0.940,0.3459)
                (0.950,0.3595)
                (0.960,0.3543)
                (0.970,0.3455)
                (0.980,0.3417)
                (0.990,0.3334)
                (1.000,0.3344)
            };
            \addplot[color=mamiColor, line width=1.5pt, solid] coordinates {
                (0.010,0.6930)
                (0.020,0.6911)
                (0.030,0.6765)
                (0.040,0.6330)
                (0.050,0.5516)
                (0.060,0.5062)
                (0.070,0.4881)
                (0.080,0.4603)
                (0.090,0.4630)
                (0.100,0.4470)
                (0.110,0.3832)
                (0.120,0.3589)
                (0.130,0.3888)
                (0.140,0.3476)
                (0.150,0.3567)
                (0.160,0.3626)
                (0.170,0.3655)
                (0.180,0.3533)
                (0.190,0.3859)
                (0.200,0.3445)
                (0.210,0.2451)
                (0.220,0.2577)
                (0.230,0.2509)
                (0.240,0.2512)
                (0.250,0.2533)
                (0.260,0.2573)
                (0.270,0.2778)
                (0.280,0.2669)
                (0.290,0.2512)
                (0.300,0.2546)
                (0.310,0.2242)
                (0.320,0.2355)
                (0.330,0.2212)
                (0.340,0.2217)
                (0.350,0.2260)
                (0.360,0.2207)
                (0.370,0.2196)
                (0.380,0.2173)
                (0.390,0.2284)
                (0.400,0.2181)
                (0.410,0.2091)
                (0.420,0.2153)
                (0.430,0.2160)
                (0.440,0.2126)
                (0.450,0.2144)
                (0.460,0.2082)
                (0.470,0.2127)
                (0.480,0.2100)
                (0.490,0.2107)
                (0.500,0.2075)
                (0.510,0.2042)
                (0.520,0.2066)
                (0.530,0.2068)
                (0.540,0.2040)
                (0.550,0.2043)
                (0.560,0.2014)
                (0.570,0.2020)
                (0.580,0.2024)
                (0.590,0.2070)
                (0.600,0.2113)
                (0.610,0.2028)
                (0.620,0.1998)
                (0.630,0.2038)
                (0.640,0.2018)
                (0.650,0.2092)
                (0.660,0.2023)
                (0.670,0.1998)
                (0.680,0.2033)
                (0.690,0.2008)
                (0.700,0.1989)
                (0.710,0.2011)
                (0.720,0.2029)
                (0.730,0.2008)
                (0.740,0.2018)
                (0.750,0.1994)
                (0.760,0.1995)
                (0.770,0.2007)
                (0.780,0.1991)
                (0.790,0.2022)
                (0.800,0.1993)
                (0.810,0.1998)
                (0.820,0.1988)
                (0.830,0.1990)
                (0.840,0.2024)
                (0.850,0.1987)
                (0.860,0.1994)
                (0.870,0.1987)
                (0.880,0.2032)
                (0.890,0.2022)
                (0.900,0.1987)
                (0.910,0.1986)
                (0.920,0.1986)
                (0.930,0.1986)
                (0.940,0.1986)
                (0.950,0.1986)
                (0.960,0.1986)
                (0.970,0.1987)
                (0.980,0.1986)
                (0.990,0.1986)
                (1.000,0.1986)
            };
            \end{axis}
        \end{tikzpicture}
    \end{subfigure}
    \caption{\textbf{Training loss evolution of \textsc{MemeWeaver} with and without IMGR across datasets.} Loss is plotted over normalized global training steps.}
    \label{fig:loss_evolution}
\end{figure}
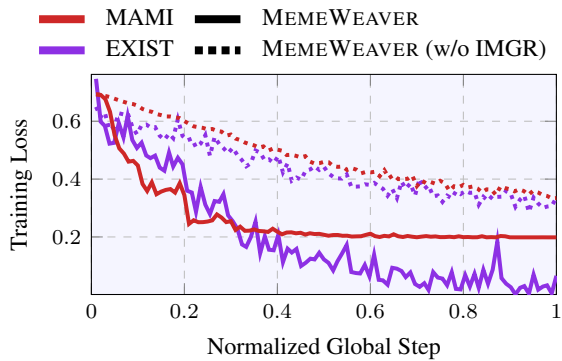

\paragraph{Training Convergence}
Beyond boosting performance, the IMGR module substantially improves CLIP's learning dynamics.
As shown in Figure~\ref{fig:loss_evolution}, the standard CLIP (dash lines) exhibits a roughly linear, gradual decrease in training loss over the full schedule.
In contrast, \textsc{MemeWeaver} plunges almost immediately---especially on MAMI---halving its loss in under 20\% of the training steps and reaching a stable plateau by 40\%.
Although the effect is less pronounced on the EXIST corpus, it still yields a faster descent and a lower steady-state loss than the vanilla model.
This faster convergence also translates into reduced computational cost and energy consumption, making \textsc{MemeWeaver} a more environmentally sustainable solution---an increasingly important criterion that should be weighed alongside accuracy in model evaluation~\citep{DBLP:conf/aaai/MoroRV23, DBLP:conf/acl/RagazziIMP24}.

\paragraph{Modality Fusion Strategy}
Table~\ref{tab:modalities_fusion} shows the impact of different fusion strategies on our best configurations across both datasets.
On EXIST, naive concatenation performs surprisingly well, outperforming both MFB and GMU.
In particular, GMU performs poorly, suggesting that gated mechanisms may be unstable in low-resource settings, where dynamic reweighting can amplify noise.
In contrast, both MFB and GMU yield substantial gains on MAMI, with MFB achieving the best overall results.
These findings highlight the data-dependent nature of fusion effectiveness and stress the need for systematic evaluation, rather than assuming that higher model complexity ensures better integration.

\begin{table}[!t]
\centering
\small
\begin{adjustbox}{width=\linewidth}
\begin{tabular}{lcccccc}
\toprule
  & \multicolumn{3}{c}{\textbf{MAMI}} 
  & \multicolumn{3}{c}{\textbf{EXIST}} \\
\cmidrule(lr{0.5em}){2-4}
\cmidrule(lr{0.5em}){5-7}
\textbf{Fusion}  & \textbf{Acc} & \textbf{F1} & \textbf{AUC} 
  & \textbf{Acc} & \textbf{F1} & \textbf{AUC} \\
\midrule
Concat & 73.6 & 73.5 & 82.0
& \textbf{76.3} & \textbf{71.6} & \textbf{75.7} \\
MFB & \textbf{77.6} & \textbf{77.4} & \textbf{83.4}
& \underline{74.6} & \underline{70.5} & \underline{75.7}  \\
GMU & \underline{74.4} & \underline{74.3} & \underline{82.1}
& 71.6 & 60.3 & 73.6 \\
\bottomrule
\end{tabular}
\end{adjustbox}
\caption{\textbf{Comparison of modality fusion strategies.} Best results are in bold and second-best are underlined.}
\label{tab:modalities_fusion}
\end{table}

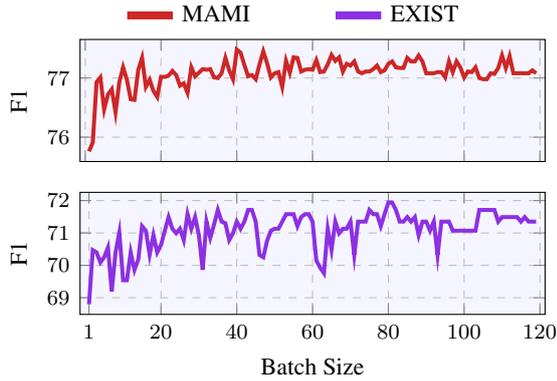
\begin{figure}[!t]
    \centering
    \begin{tikzpicture}
        \begin{axis}[
            hide axis,
            xmin=0, xmax=1, ymin=0, ymax=1,
            legend columns=2,
            legend style={
                at={(0.5,1.05)},
                anchor=south,
                draw=none,
                fill=none,
                /tikz/every even column/.append style={column sep=1cm},
                font=\footnotesize,
            }
        ]
        \addlegendimage{color=mamiColor, line width=3pt}
        \addlegendentry{MAMI}
        \addlegendimage{color=existColor, line width=3pt}
        \addlegendentry{EXIST}
        \end{axis}
    \end{tikzpicture}

    \begin{subfigure}[H]{\columnwidth}
        \begin{tikzpicture}
            \begin{axis}[
                width=\textwidth,
                height=3.2cm,
                ymajorgrids=true,
                grid=both,
                grid style=dashed,
                ytick={76, 77},
                xticklabels={},
                axis background/.style={fill=blue!4},
                enlarge x limits=0.02,
                enlarge y limits=0.1,
                ylabel={F1},
                every tick label/.append style={font=\fontsize{8}{8}\selectfont},
                xlabel style={font=\footnotesize},
    ylabel style={font=\footnotesize},
            ]
            \addplot[color=mamiColor, line width=1.5pt] coordinates {
               (1,75.7600)
                (2,75.9100)
                (3,76.9300)
                (4,77.0100)
                (5,76.4700)
                (6,76.7200)
                (7,76.8100)
                (8,76.4200)
                (9,76.9300)
                (10,77.1800)
                (11,76.9700)
                (12,76.6400)
                (13,76.6300)
                (14,77.1400)
                (15,77.3400)
                (16,76.8500)
                (17,76.9700)
                (18,76.7800)
                (19,76.6800)
                (20,77.0200)
                (21,77.0200)
                (22,77.0400)
                (23,77.1200)
                (24,77.0500)
                (25,76.8700)
                (26,76.9100)
                (27,76.7500)
                (28,77.1800)
                (29,77.0300)
                (30,77.0900)
                (31,77.1500)
                (32,77.1400)
                (33,77.1500)
                (34,77.0100)
                (35,77.0000)
                (36,77.0800)
                (37,77.3400)
                (38,77.1300)
                (39,76.9800)
                (40,77.4800)
                (41,77.4300)
                (41,77.4300)
                (42,77.2200)
                (43,77.0300)
                (44,77.0500)
                (45,76.9800)
                (46,77.2300)
                (47,77.4500)
                (48,77.2500)
                (49,77.0100)
                (50,77.0800)
                (51,77.1000)
                (52,76.8500)
                (53,77.2900)
                (54,77.0900)
                (55,77.3500)
                (56,77.3400)
                (57,77.1100)
                (58,77.1400)
                (59,77.1500)
                (60,77.1400)
                (61,77.3000)
                (62,77.1000)
                (63,77.1100)
                (64,77.2300)
                (65,77.3900)
                (66,77.2900)
                (67,77.3500)
                (68,77.1100)
                (69,77.2400)
                (70,77.2100)
                (71,77.2800)
                (72,77.1100)
                (73,77.0800)
                (74,77.1100)
                (75,77.1100)
                (76,77.1500)
                (77,77.2100)
                (78,77.1000)
                (79,77.1300)
                (80,77.2400)
                (81,77.2800)
                (82,77.1800)
                (83,77.1700)
                (84,77.1700)
                (85,77.3400)
                (86,77.2800)
                (87,77.2800)
                (88,77.3800)
                (89,77.2800)
                (90,77.0800)
                (91,77.0800)
                (92,77.0800)
                (93,77.1000)
                (94,77.1000)
                (95,77.0100)
                (96,77.2800)
                (97,77.1000)
                (98,77.2800)
                (99,77.1800)
                (100,77.0100)
                (101,77.1100)
                (102,77.1000)
                (103,77.1800)
                (104,77.0000)
                (105,76.9800)
                (106,76.9800)
                (107,77.0800)
                (108,77.0700)
                (109,77.1700)
                (110,77.3800)
                (111,77.1700)
                (112,77.3700)
                (113,77.0800)
                (114,77.0800)
                (115,77.0800)
                (116,77.0800)
                (117,77.0800)
                (118,77.1300)
                (119,77.0800)
            };
            \end{axis}
        \end{tikzpicture}
    \end{subfigure}

    \begin{subfigure}[H]{\columnwidth}
        \begin{tikzpicture}
            \begin{axis}[
                width=\textwidth,
                height=3.2cm,
                ymajorgrids=true,
                grid=both,
                xtick={1,20,40,60,80,100,120},
                grid style=dashed,
                axis background/.style={fill=blue!4},
                enlarge x limits=0.02,
                enlarge y limits=0.1,
                xlabel={Batch Size},
                ylabel={F1},
                every tick label/.append style={font=\fontsize{8}{8}\selectfont},
                xlabel style={font=\footnotesize},
    ylabel style={font=\footnotesize},
            ]
            \addplot[color=existColor, line width=1.5pt] coordinates {
                (1,68.8000)
                (2,70.4700)
                (3,70.4000)
                (4,70.0900)
                (5,70.2500)
                (6,70.5500)
                (7,69.2000)
                (8,70.4000)
                (9,71.0100)
                (10,69.5400)
                (11,69.5400)
                (12,70.4000)
                (13,69.8600)
                (14,70.1800)
                (15,71.2100)
                (16,71.0700)
                (17,70.3300)
                (18,70.9300)
                (19,70.4000)
                (20,70.6300)
                (21,71.0700)
                (22,71.4900)
                (23,71.1300)
                (24,70.9900)
                (25,71.1300)
                (26,70.7800)
                (27,71.5800)
                (28,71.1300)
                (29,71.4900)
                (30,70.9300)
                (31,69.8600)
                (32,71.3500)
                (33,70.8500)
                (34,71.2600)
                (35,71.7100)
                (36,71.3500)
                (37,71.3000)
                (38,70.7800)
                (39,70.9900)
                (40,71.3500)
                (41,71.1300)
                (42,71.3500)
                (43,71.7100)
                (44,71.7100)
                (45,71.3500)
                (46,70.3200)
                (47,70.2500)
                (48,70.7600)
                (49,71.0700)
                (50,71.1300)
                (51,71.1300)
                (52,71.3500)
                (53,71.5800)
                (54,71.5800)
                (55,71.5800)
                (56,71.2100)
                (57,71.4900)
                (58,71.5800)
                (59,71.5800)
                (60,71.3500)
                (61,70.1500)
                (62,69.9300)
                (63,69.7600)
                (64,71.1300)
                (65,70.4800)
                (66,71.3500)
                (67,70.9900)
                (68,71.1300)
                (69,71.1300)
                (70,71.3500)
                (71,70.3200)
                (72,71.3500)
                (73,71.3500)
                (74,71.3500)
                (75,71.7100)
                (76,71.5800)
                (77,71.5800)
                (78,71.2100)
                (79,71.5800)
                (80,71.9400)
                (80,71.9400)
                (81,71.9400)
                (82,71.7100)
                (83,71.3500)
                (84,71.3500)
                (85,71.4900)
                (86,71.3500)
                (87,71.4900)
                (88,70.9300)
                (89,71.3500)
                (90,71.3500)
                (91,71.0700)
                (92,71.3500)
                (93,70.3200)
                (94,71.3500)
                (95,71.3500)
                (96,71.3500)
                (97,71.0700)
                (98,71.0700)
                (99,71.0700)
                (100,71.0700)
                (101,71.0700)
                (102,71.0700)
                (103,71.0700)
                (104,71.7100)
                (105,71.7100)
                (106,71.7100)
                (107,71.7100)
                (108,71.7100)
                (109,71.3500)
                (110,71.4900)
                (111,71.4900)
                (112,71.4900)
                (113,71.4900)
                (114,71.4900)
                (115,71.3500)
                (116,71.4900)
                (117,71.3500)
                (118,71.3500)
                (119,71.3500)
            };
            \end{axis}
        \end{tikzpicture}
    \end{subfigure}
    \caption{F1 scores evolution as inference batch size increases for MAMI (top) and EXIST (bottom).}
    \label{fig:batchsize_evolution}
\end{figure}

\paragraph{Inference Batch Size}
Batch size is a key hyperparameter, as it determines graph depth and the number of inter-instance relations.
Figure~\ref{fig:batchsize_evolution} shows that F1 improves with larger inference-time batches, plateauing around 20.
On MAMI, performance remains stable beyond this point, while on EXIST it fluctuates more, likely due to the smaller training set and greater sensitivity to this hyperparameter.
However, the model stays competitive even with small batches: with size one, performance drops only slightly, especially on MAMI, despite the absence of inter-instance connections.
We attribute this robustness to training, where \textsc{MemeWeaver} reaches a much lower loss minimum than when trained without IMGR (see Figure~\ref{fig:loss_evolution}).
Thus, even when reasoning cannot be fully exploited at inference time, the model continues to benefit from its stronger training dynamics.

\paragraph{Embedding Space Analysis}
To better understand the contribution of IMGR, we compare embeddings $\mathbf{f}$ (without IMGR) and $\mathbf{f}'$ (with IMGR).
We first apply PCA~\citep{abdi2010principal} to reduce embeddings to 50 dimensions, followed by t-SNE~\citep{maaten2008visualizing} for 2D projection.
As shown in Figure~\ref{fig:emb_comparison}, the $\mathbf{f}'$ embeddings exhibit clearer separation w.r.t. ground-truth labels.
To quantify this effect, we train a linear logistic classifier on the 2D PCA$\to$t-SNE embeddings.
The 5-fold cross-validation accuracies, reported below each plot in Figure~\ref{fig:emb_comparison}, confirm the superior discriminative power of $\mathbf{f}'$, supporting our hypothesis and explaining the performance gains of our variant.

\paragraph{Embedding Affinity Analysis}
To better understand the connections learned by IMGR, we examined the relationship between meme similarity and affinity scores $R$.
Using fused multimodal embeddings $\mathbf{f}$, we computed pairwise cosine similarities across memes and grouped them by class alignment (both misogynistic/sexist, both non-misogynistic/non-sexist, or mixed).
We deliberately avoided the refined embeddings $\mathbf{f}'$, as each incorporates information from all others, which would distort pairwise similarity estimates.

\begin{figure}[!t]
    \centering
\begin{subfigure}[b]{\linewidth}
    \centering
    \includegraphics[width=\linewidth]{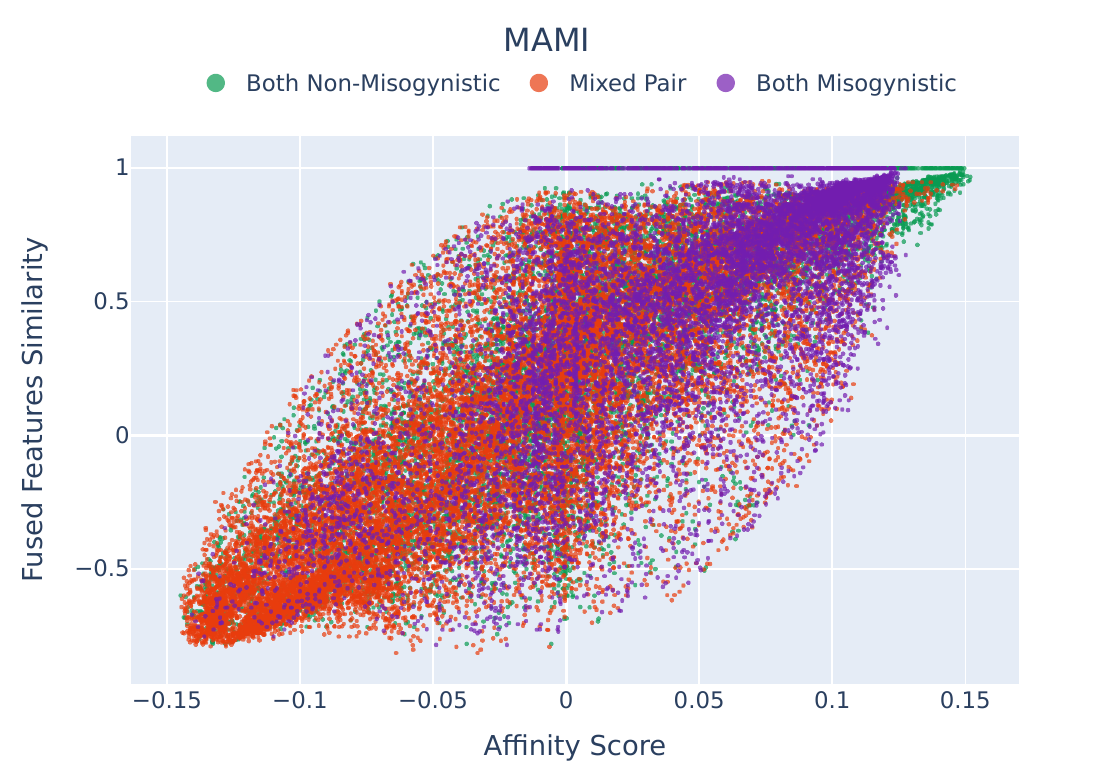}
\end{subfigure}
\begin{subfigure}[b]{\linewidth}
    \centering
    \includegraphics[width=\linewidth]{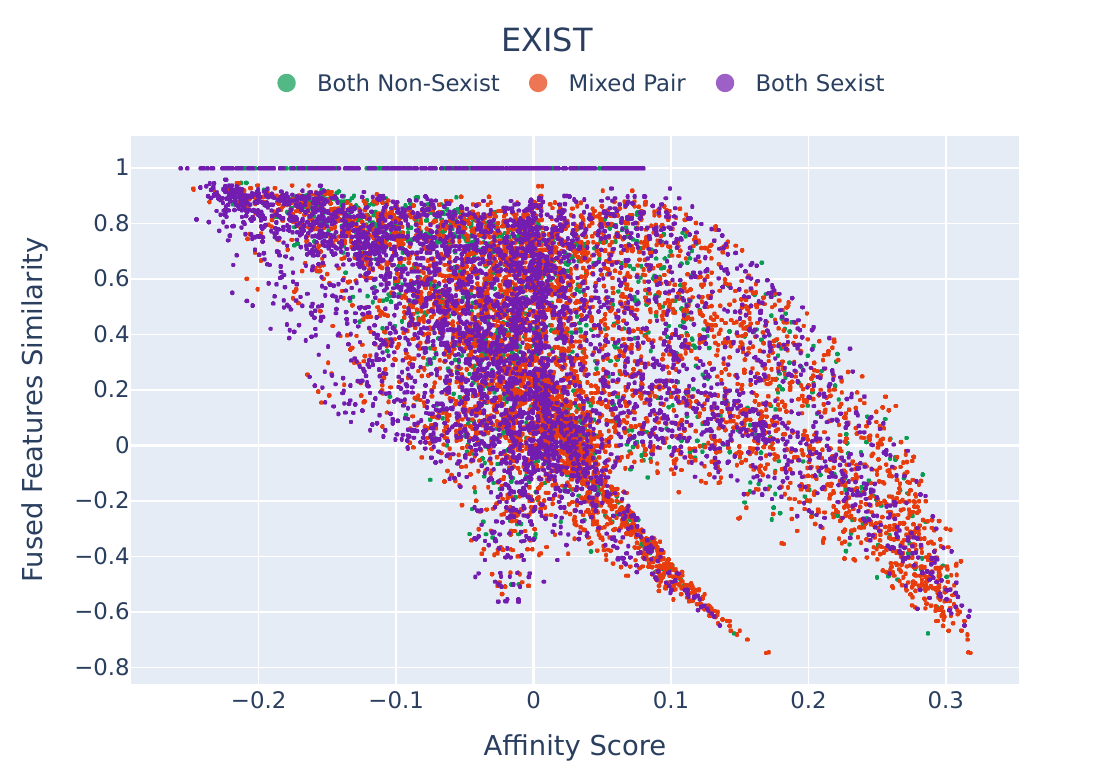} 
\end{subfigure}
\caption{\textbf{Affinity patterns in IMGR.} Correlation between learned affinity scores and fused embedding similarity, with dataset-specific behaviors.}
\label{fig:affinity_analysis}
\end{figure}

As shown in Figure~\ref{fig:affinity_analysis}, MAMI exhibits a strong positive Pearson correlation ($r=0.819$): similar memes tend to receive stronger affinity scores, reinforcing their connections.
This contrasts with the heuristic of~\citet{xu2025hyperhateprompt}, which links dissimilar instances to maximize contrastive learning.
EXIST, instead, aligns more closely with that heuristic, though with a weaker negative correlation ($r=-0.5937$).
A post hoc test further confirmed the importance of these learned patterns: simply inverting affinity signs at inference time caused a severe performance drop ($\approx$30 points), suggesting that graph structures must adapt to dataset characteristics rather than fixed heuristics.

We also find that positive pairs cluster in the upper half of the figure, indicating thematic coherence within classes and reflecting the social interplay.
Finally, both datasets contain many meme pairs with affinity scores near zero, producing a sparse graph that avoids over-connecting unrelated memes and preserves meaningful distinctions.

\paragraph{Backbone Variant}

Beyond CLIP, we explore a custom backbone for \textsc{MemeWeaver} using XLM-RoBERTa~\citep{DBLP:conf/acl/ConneauKGCWGGOZ20} as the text encoder and Vision Transformer~\citep{DBLP:conf/iclr/DosovitskiyB0WZ21} as the image encoder.
Table~\ref{tab:results_variant} compares the XLM-R/ViT variant to our best-performing CLIP-based settings (no prompts for MAMI and Prompt A for EXIST; see Table~\ref{tab:main_results}).
Although the custom backbone does not surpass the CLIP-based model, \textsc{MemeWeaver} continues to outperform standard baselines on most metrics (with the exception of F1 on EXIST), highlighting the framework's robustness and adaptability across architectures.

\begin{table}[!t]
  \centering
  \small
  \begin{adjustbox}{width=\columnwidth,center}
    \begin{tabular}{lccccccc}
      \toprule
        \multirow{2}{*}[-2pt]{\textbf{Backbone}} & \multirow{2}{*}[-2pt]{\textbf{\begin{tabular}[c]{@{}c@{}}\textsc{Meme}\\\textsc{Weaver}\end{tabular}}} & \multicolumn{3}{c}{\textbf{MAMI}} 
        & \multicolumn{3}{c}{\textbf{EXIST}} \\
      \cmidrule(lr{.5em}){3-5} \cmidrule(lr{.5em}){6-8}
        & & \textbf{Acc} & \textbf{F1} & \textbf{AUC}
        & \textbf{Acc} & \textbf{F1} & \textbf{AUC} \\
      \midrule
      \multirow{2}{*}[-1pt]{CLIP} & \xmark & 71.9 & 71.8 & 79.4 & 68.2 & 59.5 & \textbf{75.1} \\
       & \cmark & \textbf{77.6} & \textbf{77.4} & \textbf{83.4} & \textbf{76.3} & \textbf{73.6} & 72.9 \\
      \midrule
      \multirow{2}{*}[-1pt]{XLM-R/ViT} & \xmark & 66.8 & 66.8 & 74.7 & 71.1 & \textbf{66.1} & 67.2 \\
      & \cmark & \textbf{72.5} & \textbf{72.4} & \textbf{78.6} & \textbf{72.5} & 63.5 & \textbf{71.4} \\
      \bottomrule
    \end{tabular}
  \end{adjustbox}
  \caption{\textbf{Performance of \textsc{MemeWeaver} with different backbone architectures.} Best results are in bold.}
  \label{tab:results_variant}
\end{table}

\section{Conclusion}
We introduced \textsc{MemeWeaver}, a fully learnable graph-based framework for multimodal detection of sexism and misogyny in memes.
Our method consistently outperforms the state of the art on challenging benchmarks, highlighting the benefits of batch-level graph-based reasoning and advanced multimodal fusion.
Analyses of embedding spaces and affinity patterns further revealed meaningful structures and dataset-dependent behaviors, stressing the need to adapt modeling assumptions to context.
These results open avenues for more nuanced relation modeling and fine-grained archetype categorization in multimodal hate speech detection~\citep{rizzi2023recognizing}, as well as extending \textsc{MemeWeaver} to broader hate detection tasks beyond sexism and misogyny.

\section*{Limitations}
Despite its contributions, our work has several limitations that warrant further exploration.

First, while we experimented with LLMs for image captioning, we did not integrate them as backbones for \textsc{MemeWeaver}, which could enhance semantic and contextual understanding; yet this was infeasible due to computational limits, particularly constraints on batch size, which is a critical hyperparameter of our graph-based framework.

Second, we evaluated multimodal LLMs in a zero-shot setting, which is also their most common usage scenario; however, fine-tuned comparisons remain an interesting avenue for future research.

Third, we only tested two prompt types for captioning (surface-level vs. semantic inference) and did not explore other prompts or reasoning chains, which may improve visual--textual grounding.

Fourth, for EXIST-Spanish, we relied on machine translation, which we manually verified and found reliable; however, translation may still lose subtle cultural or linguistic cues, and future research should employ native multilingual encoders.

Finally, the inter-meme graph is built within fixed training batches, which ensures efficiency but may restrict the capture of global structures or cross-batch relations; future work should investigate scalable or memory-augmented graph reasoning to improve global coherence.

\section*{Ethical Considerations}

While \textsc{MemeWeaver} advances automated detection of harmful online content targeting women, the deployment of such systems raises concerns about potential biases  that may disproportionately flag content from marginalized communities or fail to capture culturally-specific forms of sexism and misogyny. The use of automated moderation systems must be balanced with human oversight, as misclassification can lead to unjust content removal or allow harmful content to remain online.

\section*{Acknowledgments}
The work of Gómez was partially supported through Grant CIACIF/2021/295 funded by GVA.
The work of Rosso was in the framework of the ANNOTATE-MULTI2 project (PID2024-156022OB-C32) funded by MICIU/AEI/10.13039/501100011033 and by ERDF/EU.
The work of Italiani, Ragazzi, and Moro was partially supported by AI-PACT project (CUP B47H22004450008, B47H22004460001); National Plan PNC-I.1 DARE initiative (PNC0000002, CUP B53C22006450001); PNRR Extended Partnership FAIR (PE00000013, Spoke 8); 2024 Scientific Research and High Technology Program, project ``AI analysis for risk assessment of empty lymph nodes in endometrial cancer surgery'', the Fondazione Cassa di Risparmio in Bologna; Chips JU TRISTAN project (G.A. 101095947).
We extend ours thanks to the Maggioli Group\footnote{\url{https://www.maggioli.com/who-we-are/company-profile}} for partially supporting the Ph.D. scholarship granted to Italiani.

\bibliography{main}

@book{herz2012content,
  title={{The Content and Context of Hate Speech: Rethinking Regulation and Responses}},
  author={Herz, Michael and Moln{\'a}r, P{\'e}ter},
  year={2012},
  publisher={Cambridge University Press}
}

@article{walther2022online,
title = {{Social Media and Online Hate}},
journal = {Current Opinion in Psychology},
volume = {45},
pages = {101298},
year = {2022},
issn = {2352-250X},
author = {Joseph B. Walther},
}

@article{fontanella2024misogyny,
  title={{How do we study misogyny in the digital age? A systematic literature review using a computational linguistic approach}},
  author={Fontanella, Lara and Chulvi, Berta and Ignazzi, Elisa and Sarra, Annalina and Tontodimamma, Alice},
  journal={Humanities and Social Sciences Communications},
  volume={11},
  number={1},
  pages={1--15},
  year={2024},
  publisher={Palgrave}
}

@article{zhang2019graph,
  title={{Graph Convolutional Networks: A Comprehensive Review}},
  author={Zhang, Si and Tong, Hanghang and Xu, Jiejun and Maciejewski, Ross},
  journal={Computational Social Networks},
  volume={6},
  number={1},
  pages={1--23},
  year={2019},
  publisher={Springer}
}

@inproceedings{li2024text,
  title={{Text-Video Retrieval via Multi-Modal Hypergraph Networks}},
  author={Li, Qian and Su, Lixin and Zhao, Jiashu and Xia, Long and Cai, Hengyi and Cheng, Suqi and Tang, Hengzhu and Wang, Junfeng and Yin, Dawei},
  booktitle={Proc. of the ACM ICWSDM},
  pages={369--377},
  year={2024}
}

@inproceedings{mafla2021multi,
  title={{Multi-Modal Reasoning Graph for Scene-Text based Fine-Grained Image Classification and Retrieval}},
  author={Mafla, Andres and Dey, Sounak and Biten, Ali Furkan and Gomez, Lluis and Karatzas, Dimosthenis},
  booktitle={Proc. of the IEEE/CVF WACV},
  pages={4023--4033},
  year={2021}
}

@inproceedings{hebert2024multi,
  title={{Multi-Modal Discussion Transformer: Integrating Text, Images and Graph Transformers to Detect Hate Speech on Social Media}},
  author={Hebert, Liam and Sahu, Gaurav and Guo, Yuxuan and Sreenivas, Nanda Kishore and Golab, Lukasz and Cohen, Robin},
  booktitle={Proc. of AAAI},
  volume={38},
  number={20},
  pages={22096--22104},
  year={2024}
}

@article{zhu2022multimodal,
  title={{Multimodal Emotion Classification with Multi-Level Semantic Reasoning Network}},
  author={Zhu, Tong and Li, Leida and Yang, Jufeng and Zhao, Sicheng and Xiao, Xiao},
  journal={IEEE Trans. on Multimedia},
  volume={25},
  pages={6868--6880},
  year={2022},
  publisher={IEEE}
}

@article{glick2001ambivalent,
  title={{Hostile and Benevolent Sexism as Complementary Justifications for Gender Inequality}},
  author={Glick, Peter and Fiske, Susan T},
  journal={American psychologist},
  volume={56},
  number={2},
  pages={109},
  year={2001},
  publisher={American Psychological Association}
}

@misc{duggan2017online,
    author={Duggan, Maeve},
    title={{Online Harassment}},
    journal={Pew Research Center},
    year={2017}, 
    howpublished={\url{https://www.pewresearch.org/internet/wp-content/uploads/sites/9/2017/07/PI_2017.07.11_Online-Harassment_FINAL.pdf}},
}

@article{bradley2015collateral,
  title={{The Collateral Damage of Ambient Sexism: Observing Sexism Impacts Bystander Self-Esteem and Career Aspirations}},
  author={Bradley-Geist, Jill C and Rivera, Ivy and Geringer, Susan D},
  journal={Sex Roles},
  volume={73},
  pages={29--42},
  year={2015},
  publisher={Springer}
}

@article{kowalski2015cyberbullying,
  title={{Cyberbullying: Prevalence, Causes, and Consequences}},
  author={Kowalski, Robin M and Whittaker, Elizabeth},
  journal={The Wiley handbook of Psychology, Technology, and Society},
  pages={142--157},
  year={2015},
  publisher={Wiley Online Library}
}

@inproceedings{gomez2020exploring,
  title={{Exploring Hate Speech Detection in Multimodal Publications}},
  author={Gomez, Raul and Gibert, Jaume and Gomez, Lluis and Karatzas, Dimosthenis},
  booktitle={Proc. of the IEEE/CVF WACV},
  pages={1470--1478},
  year={2020},
  doi={10.1109/WACV45572.2020.9093414},
}

@inproceedings{cocchieri-etal-2025-call,
    title = "``What do you call a dog that is incontrovertibly true? Dogma'': Testing {LLM} Generalization through Humor",
    author = "Cocchieri, Alessio  and
      Ragazzi, Luca  and
      Italiani, Paolo  and
      Tagliavini, Giuseppe  and
      Moro, Gianluca",
    editor = "Che, Wanxiang  and
      Nabende, Joyce  and
      Shutova, Ekaterina  and
      Pilehvar, Mohammad Taher",
    booktitle = "Proceedings of the 63rd Annual Meeting of the Association for Computational Linguistics (Volume 1: Long Papers)",
    month = jul,
    year = "2025",
    address = "Vienna, Austria",
    publisher = "Association for Computational Linguistics",
    url = "https://aclanthology.org/2025.acl-long.1117/",
    doi = "10.18653/v1/2025.acl-long.1117",
    pages = "22922--22937",
    ISBN = "979-8-89176-251-0"
}

@inproceedings{khan2021exploiting,
  title={{Exploiting BERT for Multimodal Target Sentiment Classification through Input Space Translation}},
  author={Khan, Zaid and Fu, Yun},
  booktitle={Proc. of ACM ICM},
  pages={3034--3042},
  year={2021}
}

@inproceedings{fersini2022mami,
  author={Elisabetta Fersini and Francesca Gasparini and Giulia Rizzi and Aurora Saibene and Berta Chulvi and Paolo Rosso and Alyssa Lees and Jeffrey Sorensen},
  title={{SemEval-2022 Task 5: Multimedia Automatic Misogyny Identification}},
  booktitle={Proc. of the 16th SemEval@NAACL},
  pages={533--549},
  publisher={ACL},
  year={2022},
  doi={https://doi.org/10.18653/v1/2022.semeval-1.74},
}

@article{richardson2018woman,
  title={{Woman-Hating: On Misogyny, Sexism, and Hate Speech}},
  author={Richardson-Self, Louise},
  journal={Hypatia},
  volume={33},
  number={2},
  pages={256--272},
  year={2018},
  publisher={Cambridge University Press}
}

@article{rizzi2024pink,
  title={{PINK at EXIST2024: A Cross-Lingual and Multi-Modal Transformer Approach for Sexism Detection in Memes}},
  author={Rizzi, Giulia and Gimeno-G{\'o}mez, David and Fersini, Elisabetta and Mart{\'\i}nez-Hinarejos, Carlos-D},
  journal={Working Notes of CLEF},
  year={2024}
}

@article{MORO2023126356,
    title = {Align-then-abstract representation learning for low-resource summarization},
    journal = {Neurocomputing},
    volume = {548},
    pages = {126356},
    year = {2023},
    issn = {0925-2312},
    doi = {https://doi.org/10.1016/j.neucom.2023.126356},
    url = {https://www.sciencedirect.com/science/article/pii/S0925231223004794},
    author = {Gianluca Moro and Luca Ragazzi}
}

@inproceedings{plaza2025exist,
  author={Laura Plaza and Jorge Carrillo{-}de{-}Albornoz and Iv{\'{a}}n {\'{A}}rcos and Paolo Rosso and Damiano Spina and Enrique Amig{\'{o}} and Julio Gonzalo and Roser Morante},
  title={{EXIST 2025: Learning with Disagreement for Sexism Identification and Characterization in Tweets, Memes, and TikTok Videos}},
  booktitle={Proc. of the 47th ECIR},
  series={Lecture Notes in Computer Science},
  volume={15576},
  pages={442--449},
  publisher={Springer},
  year={2025},
  doi={10.1007/978-3-031-88720-8\_65}
}

@article{hodson2010joke,
  title={{A Joke is Just a Joke (Except When It Isn't): Cavalier Humor Beliefs Facilitate the Expression of Group Dominance Motives}},
  author={Hodson, Gordon and Rush, Jonathan and MacInnis, Cara C},
  journal={Journal of personality and social psychology},
  volume={99},
  number={4},
  pages={660},
  year={2010},
  publisher={American Psychological Association}
}

@article{Moro_Ragazzi_2022, 
    title={Semantic Self-Segmentation for Abstractive Summarization of Long Documents in Low-Resource Regimes}, 
    volume={36}, 
    url={https://ojs.aaai.org/index.php/AAAI/article/view/21357}, 
    DOI={10.1609/aaai.v36i10.21357},
    number={10}, 
    journal={Proceedings of the AAAI Conference on Artificial Intelligence}, author={Moro, Gianluca and Ragazzi, Luca}, 
    year={2022}, 
    month={Jun.}, 
    pages={11085-11093} 
}

@article{drucker2014sarcasm,
  title={{On Sarcasm, Social Awareness, and Gender}},
  author={Drucker, Ari and Fein, Ofer and Bergerbest, Dafna and Giora, Rachel},
  journal={Humor},
  volume={27},
  number={4},
  pages={551--573},
  year={2014},
  publisher={De Gruyter Mouton}
}

@inproceedings{plaza2024overview2,
  author={Plaza, Laura  and Carrillo{-}de{-}Albornoz, Jorge  and Ruiz, V\'{i}ctor  and Maeso, Alba  and Chulvi, Berta  and Rosso, Paolo  and Amig\'{o}, Enrique  and Gonzalo, Julio  and Morante, Roser  and Spina, Damiano},
  title={{Overview of EXIST 2024 -- Learning with Disagreement for Sexism Identification and Characterization in Social Networks and Memes (Extended Overview)}},
  booktitle={Working Notes of {CLEF}},
  year={2024},
}

@inproceedings{jha2017does,
  title={{When Does a Compliment Become Sexist? Analysis and Classification of Ambivalent Sexism using Twitter Data}},
  author={Jha, Akshita and Mamidi, Radhika},
  booktitle={Proc. of the Workshop on NLP and Computational Social Science},
  pages={7--16},
  year={2017}
}

@inproceedings{anzovino2018automatic,
  title={{Automatic Identification and Classification of Misogynistic Language on Twitter}},
  author={Anzovino, Maria and Fersini, Elisabetta and Rosso, Paolo},
  booktitle={Proc. of NLDB},
  pages={57--64},
  year={2018},
  organization={Springer}
}

@article{parikh2021categorizing,
  title={{Categorizing Sexism and Misogyny through Neural Approaches}},
  author={Parikh, Pulkit and Abburi, Harika and Chhaya, Niyati and Gupta, Manish and Varma, Vasudeva},
  journal={ACM Trans. on Web},
  volume={15},
  number={4},
  pages={1--31},
  year={2021},
  publisher={ACM New York, NY}
}

@inproceedings{de2023ai,
  title={{AI-UPV at EXIST 2023--Sexism Characterization Using Large Language Models Under The Learning with Disagreements Regime}},
  author={de Paula, A and Rizzi, G and Fersini, E and Spina, D},
  booktitle={CEUR WORKSHOP PROCEEDINGS},
  volume={3497},
  pages={985--999},
  year={2023},
  organization={CEUR-WS}
}

@inproceedings{xu2025hyperhateprompt,
    title={{HyperHatePrompt: A Hypergraph-based Prompting Fusion Model for Multimodal Hate Detection}},
    author={Xu, Bo  and  Yu, Erchen  and  Zhou, Jiahui  and  Lin, Hongfei  and 
 Zong, Linlin},
    booktitle={Proc. of the 31st ICCL},
    year={2025},
    publisher={ACL},
    pages={3825--3835},
}

@inproceedings{samani2025large,
  title={{Large Language Models with Reinforcement Learning from Human Feedback Approach for Enhancing Explainable Sexism Detection}},
  author={Samani, Ali Riahi and Wang, Tianhao and Li, Kangshuo and Chen, Feng},
  booktitle={Proc. of the 31st ICCL},
  pages={6230--6243},
  year={2025}
}

@inproceedings{tian2023efficient,
  title={Efficient Multilingual Sexism Detection via Large Language Model Cascades.},
  author={Tian, Lin and Huang, Nannan and Zhang, Xiuzhen},
  booktitle={CLEF (Working Notes)},
  pages={1083--1090},
  year={2023}
}

@article{rehman2025context,
  title={{A Context-Aware Attention and Graph Neural Network-based Multimodal Framework for Misogyny Detection}},
  author={Rehman, Mohammad Zia Ur and Zahoor, Sufyaan and Manzoor, Areeb and Maqbool, Musharaf and Kumar, Nagendra},
  journal={Information Processing \& Management},
  volume={62},
  number={1},
  pages={103895},
  year={2025},
  publisher={Elsevier}
}

@inproceedings{cao2022prompting,
  title={{Prompting for Multimodal Hateful Meme Classification}},
  author={Cao, Rui and Lee, Roy Ka-Wei and Chong, Wen-Haw and Jiang, Jing},
  booktitle={Proc. of EMNLP},
  pages={321--332},
  year={2022}
}

@inproceedings{cao2023procap,
  title={{Pro-Cap: Leveraging a Frozen Vision-Language Model for Hateful Meme Detection}},
  author={Cao, Rui and Hee, Ming Shan and Kuek, Adriel and Chong, Wen-Haw and Lee, Roy Ka-Wei and Jiang, Jing},
  booktitle={Proc. of the 31st ACM ICM},
  pages={5244--5252},
  year={2023}
}

@article{rizzi2023recognizing,
  title={Recognizing Misogynous Memes: Biased Models and Tricky Archetypes},
  author={Rizzi, Giulia and Gasparini, Francesca and Saibene, Aurora and Rosso, Paolo and Fersini, Elisabetta},
  journal={Information Processing \& Management},
  volume={60},
  number={5},
  pages={103474},
  year={2023},
  publisher={Elsevier}
}

@ARTICLE{luo2025survey,
  author={Luo, Xuan and Liang, Bin and Wang, Qianlong and Li, Jing and Cambria, Erik and Zhang, Xiaojun and He, Yulan and Yang, Min and Xu, Ruifeng},
  journal={IEEE Trans. on Computational Social Systems}, 
  title={{A Literature Survey on Multimodal and Multilingual Sexism Detection}}, 
  year={2025},
  volume={},
  number={},
  pages={1-19},
  doi={10.1109/TCSS.2025.3561921}}

@inproceedings{arcos2024sexism,
  title={Sexism Identification on TikTok: A Multimodal AI Approach with Text, Audio, and Video},
  author={Arcos, Iv{\'a}n and Rosso, Paolo},
  booktitle={Proc. of CLEF},
  pages={61--73},
  year={2024},
  organization={Springer}
}

@article{ragazzietalTASLP25,
  author={Ragazzi, Luca and Moro, Gianluca and Valgimigli, Lorenzo and Fiorani, Riccardo},
  journal={IEEE Transactions on Audio, Speech and Language Processing}, 
  title={Cross-Document Distillation via Graph-Based Summarization of Extracted Essential Knowledge}, 
  year={2025},
  volume={33},
  number={},
  pages={518-527},
  doi={10.1109/TASLP.2024.3490375}
}

@inproceedings{DBLP:conf/iclr/KimOLKHZ17,
  author       = {Jin{-}Hwa Kim and
                  Kyoung Woon On and
                  Woosang Lim and
                  Jeonghee Kim and
                  Jung{-}Woo Ha and
                  Byoung{-}Tak Zhang},
  title        = {Hadamard Product for Low-rank Bilinear Pooling},
  booktitle    = {Proc. of ICLR},
  publisher    = {OpenReview.net},
  year         = {2017},
  url          = {https://openreview.net/forum?id=r1rhWnZkg},
  bibsource    = {dblp computer science bibliography, https://dblp.org}
}

@inproceedings{DBLP:conf/iclr/OvalleSMG17,
  author       = {John Edison Arevalo Ovalle and
                  Thamar Solorio and
                  Manuel Montes{-}y{-}G{\'{o}}mez and
                  Fabio A. Gonz{\'{a}}lez},
  title        = {Gated Multimodal Units for Information Fusion},
  booktitle    = {Proc. of ICLR},
  publisher    = {OpenReview.net},
  year         = {2017},
  url          = {https://openreview.net/forum?id=S12\_nquOe},
  bibsource    = {dblp computer science bibliography, https://dblp.org}
}

@inproceedings{li2019visual,
  title={{Visual Semantic Reasoning for Image-Text Matching}},
  author={Li, Kunpeng and Zhang, Yulun and Li, Kai and Li, Yuanyuan and Fu, Yun},
  booktitle={Proc. of the IEEE/CVF ICCV},
  pages={4654--4662},
  year={2019}
}

@inproceedings{DBLP:conf/iclr/MoroRVVF24,
  author       = {Gianluca Moro and
                  Luca Ragazzi and
                  Lorenzo Valgimigli and
                  Fabian Vincenzi and
                  Davide Freddi},
  title        = {Revelio: Interpretable Long-Form Question Answering},
  booktitle    = {The Second Tiny Papers Track at {ICLR} 2024, Tiny Papers @ {ICLR}
                  2024, Vienna, Austria, May 11, 2024},
  publisher    = {OpenReview.net},
  year         = {2024},
  url          = {https://openreview.net/forum?id=fyvEJXsaQf},
  bibsource    = {dblp computer science bibliography, https://dblp.org}
}

@article{moro2023multi,
  title={Multi-language transfer learning for low-resource legal case summarization},
  author={Gianluca Moro and Nicola Piscaglia and Luca Ragazzi and others},
  journal={Artificial Intelligence and Law},
url={https://doi.org/10.1007/s10506-023-09373-8},
doi={10.1007/s10506-023-09373-8},
  pages={1--29},
  year={2023},
  publisher={Springer},
}

@inproceedings{DBLP:conf/ecai/MoroRV23,
  author       = {Gianluca Moro and
                  Luca Ragazzi and
                  Lorenzo Valgimigli},
  title        = {Graph-Based Abstractive Summarization of Extracted Essential Knowledge for Low-Resource Scenarios},
  booktitle    = {{ECAI}, September 30 - October 4, 2023},
  series       = {FAIA},
  volume       = {372},
  pages        = {1747--1754},
  publisher    = {{IOS} Press},
  year         = {2023},
  url          = {https://doi.org/10.3233/FAIA230460},
  doi          = {10.3233/FAIA230460},
  bibsource    = {dblp computer science bibliography, https://dblp.org},
}

@inproceedings{DBLP:conf/aaai/MoroRV23,
  author       = {Gianluca Moro and
                  Luca Ragazzi and
                  Lorenzo Valgimigli},
  editor       = {Brian Williams and
                  Yiling Chen and
                  Jennifer Neville},
  title        = {Carburacy: Summarization Models Tuning and Comparison in Eco-Sustainable
                  Regimes with a Novel Carbon-Aware Accuracy},
  booktitle    = {Thirty-Seventh {AAAI} Conference on Artificial Intelligence, {AAAI}
                  2023, Thirty-Fifth Conference on Innovative Applications of Artificial
                  Intelligence, {IAAI} 2023, Thirteenth Symposium on Educational Advances
                  in Artificial Intelligence, {EAAI} 2023, Washington, DC, USA, February
                  7-14, 2023},
  pages        = {14417--14425},
  publisher    = {{AAAI} Press},
  year         = {2023},
  url          = {https://doi.org/10.1609/aaai.v37i12.26686},
  doi          = {10.1609/AAAI.V37I12.26686},
  bibsource    = {dblp computer science bibliography, https://dblp.org}
}

@inproceedings{radford2021learning,
  title={Learning transferable visual models from natural language supervision},
  author={Radford, Alec and Kim, Jong Wook and Hallacy, Chris and Ramesh, Aditya and Goh, Gabriel and Agarwal, Sandhini and Sastry, Girish and Askell, Amanda and Mishkin, Pamela and Clark, Jack and others},
  booktitle={International conference on machine learning},
  pages={8748--8763},
  year={2021},
  organization={PmLR}
}

@inproceedings{koehn2004statistical,
  title={Statistical significance tests for machine translation evaluation},
  author={Koehn, Philipp},
  booktitle={Proceedings of the 2004 conference on empirical methods in natural language processing},
  pages={388--395},
  year={2004}
}

@article{bai2025qwen2,
  title={Qwen2. 5-vl technical report},
  author={Bai, Shuai and Chen, Keqin and Liu, Xuejing and Wang, Jialin and Ge, Wenbin and Song, Sibo and Dang, Kai and Wang, Peng and Wang, Shijie and Tang, Jun and others},
  journal={arXiv preprint arXiv:2502.13923},
  year={2025}
}

@article{abdin2024phi,
  title={Phi-4 technical report},
  author={Abdin, Marah and Aneja, Jyoti and Behl, Harkirat and Bubeck, S{\'e}bastien and Eldan, Ronen and Gunasekar, Suriya and Harrison, Michael and Hewett, Russell J and Javaheripi, Mojan and Kauffmann, Piero and others},
  journal={arXiv preprint arXiv:2412.08905},
  year={2024}
}

@article{hurst2024gpt,
  title={Gpt-4o system card},
  author={Hurst, Aaron and Lerer, Adam and Goucher, Adam P and Perelman, Adam and Ramesh, Aditya and Clark, Aidan and Ostrow, AJ and Welihinda, Akila and Hayes, Alan and Radford, Alec and others},
  journal={arXiv preprint arXiv:2410.21276},
  year={2024}
}

@article{abdi2010principal,
  title={Principal component analysis},
  author={Abdi, Herv{\'e} and Williams, Lynne J},
  journal={Wiley interdisciplinary reviews: computational statistics},
  volume={2},
  number={4},
  pages={433--459},
  year={2010},
  publisher={Wiley Online Library}
}

@article{maaten2008visualizing,
  title={Visualizing data using t-SNE},
  author={Maaten, Laurens van der and Hinton, Geoffrey},
  journal={Journal of machine learning research},
  volume={9},
  number={Nov},
  pages={2579--2605},
  year={2008}
}

@inproceedings{DBLP:conf/acl/ConneauKGCWGGOZ20,
  author       = {Alexis Conneau and
                  Kartikay Khandelwal and
                  Naman Goyal and
                  Vishrav Chaudhary and
                  Guillaume Wenzek and
                  Francisco Guzm{\'{a}}n and
                  Edouard Grave and
                  Myle Ott and
                  Luke Zettlemoyer and
                  Veselin Stoyanov},
  editor       = {Dan Jurafsky and
                  Joyce Chai and
                  Natalie Schluter and
                  Joel R. Tetreault},
  title        = {Unsupervised Cross-lingual Representation Learning at Scale},
  booktitle    = {Proceedings of the 58th Annual Meeting of the Association for Computational
                  Linguistics, {ACL} 2020, Online, July 5-10, 2020},
  pages        = {8440--8451},
  publisher    = {Association for Computational Linguistics},
  year         = {2020},
  url          = {https://doi.org/10.18653/v1/2020.acl-main.747},
  doi          = {10.18653/V1/2020.ACL-MAIN.747},
  bibsource    = {dblp computer science bibliography, https://dblp.org}
}

@inproceedings{DBLP:conf/iclr/DosovitskiyB0WZ21,
  author       = {Alexey Dosovitskiy and
                  Lucas Beyer and
                  Alexander Kolesnikov and
                  Dirk Weissenborn and
                  Xiaohua Zhai and
                  Thomas Unterthiner and
                  Mostafa Dehghani and
                  Matthias Minderer and
                  Georg Heigold and
                  Sylvain Gelly and
                  Jakob Uszkoreit and
                  Neil Houlsby},
  title        = {An Image is Worth 16x16 Words: Transformers for Image Recognition
                  at Scale},
  booktitle    = {9th International Conference on Learning Representations, {ICLR} 2021,
                  Virtual Event, Austria, May 3-7, 2021},
  publisher    = {OpenReview.net},
  year         = {2021},
  url          = {https://openreview.net/forum?id=YicbFdNTTy},
  bibsource    = {dblp computer science bibliography, https://dblp.org}
}

@article{khoong2019assessing,
  title   = {Assessing the Use of Google Translate for Spanish and Chinese Translations of Emergency Department Discharge Instructions},
  author  = {Khoong, Elizabeth C. and Steinbrook, Eric and Brown, Catherine and Fernandez, Alicia},
  journal = {JAMA Internal Medicine},
  volume  = {179},
  number  = {4},
  pages   = {580--582},
  year    = {2019},
  doi     = {10.1001/jamainternmed.2018.7653},
  pmid    = {30801626},
  pmcid   = {PMC6450297}
}

@article{taira2021pragmatic,
  title   = {A Pragmatic Assessment of Google Translate for Emergency Department Instructions},
  author  = {Taira, Breena R. and Kreger, Vanessa and Orue, Andrea and Diamond, Lisa C.},
  journal = {Journal of General Internal Medicine},
  volume  = {36},
  number  = {11},
  pages   = {3361--3365},
  year    = {2021},
  doi     = {10.1007/s11606-021-06666-z},
  pmid    = {33674922},
  pmcid   = {PMC8606479}
}

@inproceedings{zhu-etal-2019-ncls,
    title = "{NCLS}: Neural Cross-Lingual Summarization",
    author = "Zhu, Junnan  and
      Wang, Qian  and
      Wang, Yining  and
      Zhou, Yu  and
      others",
    booktitle = {{EMNLP-IJCNLP}},
    month = nov,
    year = "2019",
    address = "Hong Kong, China",
    publisher = {{ACL}},
    url = "https://aclanthology.org/D19-1302",
    doi = "10.18653/v1/D19-1302",
    pages = "3054--3064",
}

@inproceedings{DBLP:conf/acl/RagazziIMP24,
  author       = {Luca Ragazzi and
                  Paolo Italiani and
                  Gianluca Moro and
                  Mattia Panni},
  editor       = {Lun{-}Wei Ku and
                  Andre Martins and
                  Vivek Srikumar},
  title        = {What Are You Token About? Differentiable Perturbed Top-k Token Selection
                  for Scientific Document Summarization},
  booktitle    = {Findings of the Association for Computational Linguistics, {ACL} 2024,
                  Bangkok, Thailand and virtual meeting, August 11-16, 2024},
  pages        = {9427--9440},
  publisher    = {Association for Computational Linguistics},
  year         = {2024},
  url          = {https://doi.org/10.18653/v1/2024.findings-acl.561},
  doi          = {10.18653/V1/2024.FINDINGS-ACL.561},
  bibsource    = {dblp computer science bibliography, https://dblp.org}
}

@inproceedings{feng-etal-2022-msamsum,
  author       = {Xiachong Feng and
                  Xiaocheng Feng and
                  Bing Qin},
  editor       = {Song Feng and
                  Hui Wan and
                  Caixia Yuan and
                  Han Yu},
  title        = {MSAMSum: Towards Benchmarking Multi-lingual Dialogue Summarization},
  booktitle    = {DialDoc@ACL 2022},
  pages        = {1--12},
  publisher    = {{ACL}},
  year         = {2022},
  url          = {https://doi.org/10.18653/v1/2022.dialdoc-1.1},
  doi          = {10.18653/v1/2022.dialdoc-1.1},
  bibsource    = {dblp computer science bibliography, https://dblp.org}
}

\appendix

\section{Translation Quality}
\label{app:translation_quality}

Assessing translation quality is important for evaluating the reliability of the Spanish-to-English preprocessing step.
However, gold-standard English references are not available for the EXIST dataset, which prevents the direct evaluation for Spanish-to-English translation.
To address this limitation, we estimated translation quality using a standard \emph{back-translation} protocol~\citep{zhu-etal-2019-ncls, feng-etal-2022-msamsum}.
Specifically, the original Spanish text (Spanish$_1$) was translated into English using Google Translate, and the resulting English text was then translated back into Spanish (Spanish$_2$).
Translation quality was assessed by comparing Spanish$_1$ and Spanish$_2$ using automatic overlap-based metrics.
Using this approach, we obtained a BLEU-1 score of 57.30 and a ROUGE-1 score of 76.44.
These results indicate a high degree of lexical overlap between the original and back-translated Spanish text, suggesting that the English intermediate translation preserved most of the core semantic content.
This outcome is consistent with the characteristics of EXIST, where meme texts are typically short, informal, and syntactically simple.

Moreover, the observed translation quality is consistent with prior findings in the literature, which have consistently shown that Google Translate achieves particularly high accuracy for Spanish-to-English translation, one of its strongest-performing language pairs~\citep{khoong2019assessing, taira2021pragmatic, moro2023multi}.
Overall, these results suggest that the translation step introduces limited distortion and does not compromise the reliability of the downstream analysis.

\section{Prompts}\label{app:prompts}

\begin{figure}[!htb]
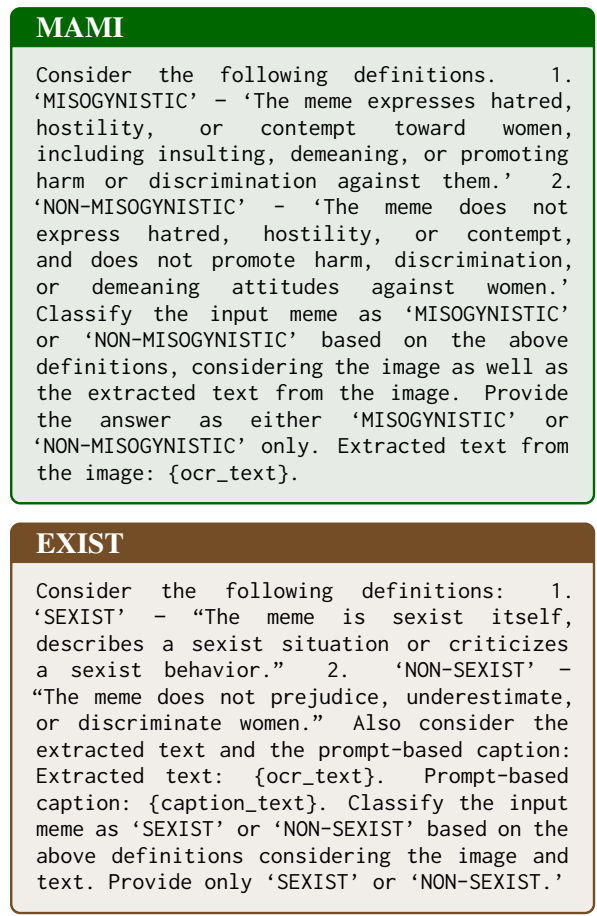

  \centering
    \begin{varwidth}{\columnwidth}
      \begin{tcolorbox}[
      colback=green!20!black!10!white,
      colframe=green!40!black,
      title=MAMI,
      fonttitle=\bfseries,
      rounded corners,        
      boxrule=1pt,
      left=6pt, right=6pt, top=4pt, bottom=4pt,
      width=\columnwidth,
      fontupper=\ttfamily\footnotesize\itshape,  
      enhanced jigsaw         
    ]
        Consider the following definitions. 1. ‘MISOGYNISTIC’ – ‘The meme expresses hatred, hostility, or contempt toward women, including insulting, demeaning, or promoting harm or discrimination against them.’  
        2. ‘NON-MISOGYNISTIC’ - ‘The meme does not express hatred, hostility, or contempt, and does not promote harm, discrimination, or demeaning attitudes against women.’  
        Classify the input meme as ‘MISOGYNISTIC’ or ‘NON-MISOGYNISTIC’ based on the above definitions, considering the image as well as the extracted text from the image. Provide the answer as either ‘MISOGYNISTIC’ or ‘NON-MISOGYNISTIC’ only.  
        Extracted text from the image: \{ocr\_text\}.
      \end{tcolorbox}


      \begin{tcolorbox}[
      colback=brown!60!black!10!white,
      colframe=brown!60!black,
      title=EXIST,
      fonttitle=\bfseries,
      rounded corners,        
      boxrule=1pt,
      left=6pt, right=6pt, top=4pt, bottom=4pt,
      width=\columnwidth,
      fontupper=\ttfamily\footnotesize\itshape,  
      enhanced jigsaw         
    ]
        Consider the following definitions:  
        1. ‘SEXIST’ – ``The meme is sexist itself, describes a sexist situation or criticizes a sexist behavior.''  
        2. ‘NON-SEXIST’ – ``The meme does not prejudice, underestimate, or discriminate women.''  
        Also consider the extracted text and the prompt‐based caption:  
        Extracted text: \{ocr\_text\}.  
        Prompt‐based caption: \{caption\_text\}.  
        Classify the input meme as ‘SEXIST’ or ‘NON-SEXIST’ based on the above definitions considering the image and text. Provide only ‘SEXIST’ or ‘NON-SEXIST.’
      \end{tcolorbox}
      \caption{Zero-shot LLM prompts across datasets.}
      \label{fig:llm_zeroshot}
    \end{varwidth}
\end{figure}

Figure~\ref{fig:llm_zeroshot} reports the prompts used for zero-shot experiments using LLMs. For the sake of reproducibility, the definitions of ``misogynistic'' and ``sexist'' included in these prompts were taken directly from the original MAMI and EXIST task guideline descriptions.

\section{LLM-based Image Captioning}\label{app:examples}

\begin{figure*}[!t]
\centering
\begin{minipage}[c]{0.3\textwidth}
    \centering
    \fbox{\includegraphics[width=\linewidth,height=0.45\textheight,keepaspectratio]{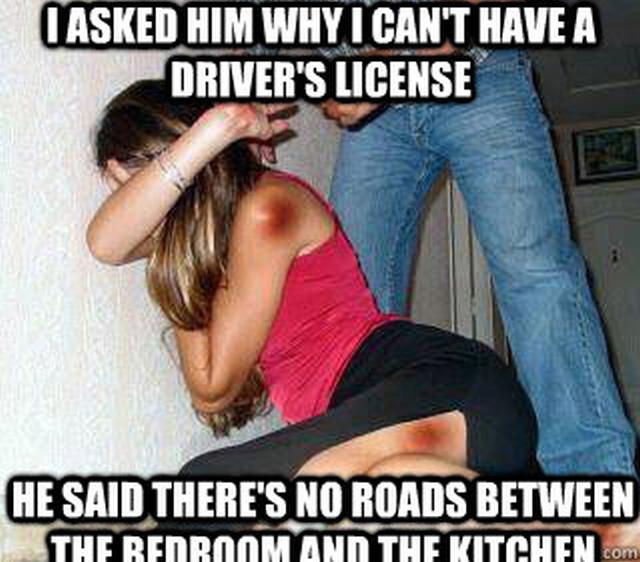}}
    \textit{\textbf{Label:} Misogynistic}
\end{minipage}
\hfill
\begin{minipage}[c]{0.65\textwidth}
    \begin{tcolorbox}[colback=white, colframe=black!10,
                      boxrule=0pt, sharp corners, 
                      valign=center, 
                      left=2pt, right=2pt, top=2pt, bottom=2pt,
                      boxsep=0pt, enhanced]
        \begin{tcolorbox}[colback=orange!80!black!10!white, colframe=orange!80!black,
                          title=Prompt A Output (Surface-level description), fonttitle=\bfseries,
                          boxrule=0.8pt, sharp corners, fontupper=\small,
                          left=5pt, right=5pt, top=5pt, bottom=5pt, rounded corners,]
        The image shows a woman in a red tank top and black pants, appearing \textbf{distressed} with her head down and hands covering her face. A person wearing blue jeans is standing behind her, holding her by the shoulders. The setting appears to be indoors, possibly a bedroom or a similar private space.
        \end{tcolorbox}

        \begin{tcolorbox}[colback=customblue2!10!white, colframe=customblue2,
                          title=Prompt B Output (Multimodal semantic inference), fonttitle=\bfseries,
                          boxrule=0.8pt, sharp corners, fontupper=\small,
                          left=5pt, right=5pt, top=5pt, bottom=5pt, rounded corners,]
        The meme contains a humorous and somewhat \textbf{crude joke} about a relationship dynamic, where one partner is being told they can't have a driver's license because there are no roads between the bedroom and the kitchen. This type of humor often relies on \textbf{stereotypes} and can be seen as \textbf{sexist}, as it implies that relationships should be confined to domestic spaces without broader societal implications. The use of such language can perpetuate gender stereotypes and limit perceptions of what relationships and personal freedom should encompass.
        \end{tcolorbox}
    \end{tcolorbox}
\end{minipage}%
\vspace{0.5cm}
\hrule
\vspace{0.5cm}
\begin{minipage}[c]{0.3\textwidth}
    \centering
    \fbox{\includegraphics[width=\linewidth,height=0.45\textheight,keepaspectratio]{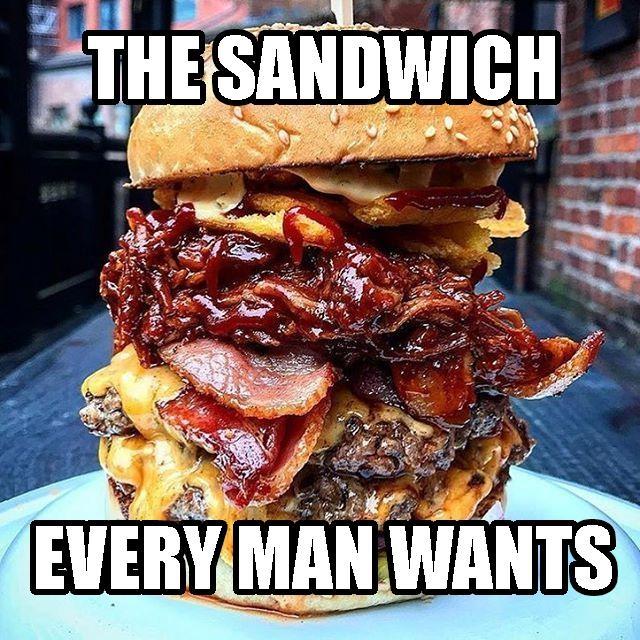}}
    \textit{\textbf{Label:} Non-Misogynistic}
\end{minipage}
\hfill
\begin{minipage}[c]{0.65\textwidth}
    \begin{tcolorbox}[colback=white, colframe=black!10,
                      boxrule=0pt, sharp corners, 
                      valign=center,
                      left=2pt, right=2pt, top=2pt, bottom=2pt,
                      boxsep=0pt, enhanced]
        \begin{tcolorbox}[colback=orange!80!black!10!white, colframe=orange!80!black,
                          title=Prompt A Output (Surface-level description), fonttitle=\bfseries,
                          boxrule=0.8pt, sharp corners, fontupper=\small,
                          left=5pt, right=5pt, top=5pt, bottom=5pt, rounded corners,]
        The image shows an extremely large sandwich stacked high with various ingredients, including what appears to be pulled meat, bacon, cheese, and possibly some sauce. The sandwich is placed on a plate, and the background suggests an outdoor setting with a brick wall and a dark surface beneath the plate. The overall presentation emphasizes the size and abundance of the sandwich.
        \end{tcolorbox}

        \begin{tcolorbox}[colback=customblue2!10!white, colframe=customblue2,
                          title=Prompt B Output (Multimodal semantic inference), fonttitle=\bfseries,
                          boxrule=0.8pt, sharp corners, fontupper=\small,
                          left=5pt, right=5pt, top=5pt, bottom=5pt, rounded corners,]
        The meme humorously exaggerates the appeal of a sandwich by labeling it as "every man wants," which could be seen as a \textbf{stereotype}. However, the intent appears to be \textbf{playful rather than derogatory}, and the use of "man" might be intended \textbf{inclusively} to refer to any person. The focus is on the sandwich's excessive layers, suggesting it's a highly desirable food item.
        \end{tcolorbox}
    \end{tcolorbox}
\end{minipage}%
\vspace{0.5cm}
\hrule
\vspace{0.5cm}
\begin{minipage}[c]{0.3\textwidth}
    \centering
    \fbox{\includegraphics[width=\linewidth,height=0.45\textheight,keepaspectratio]{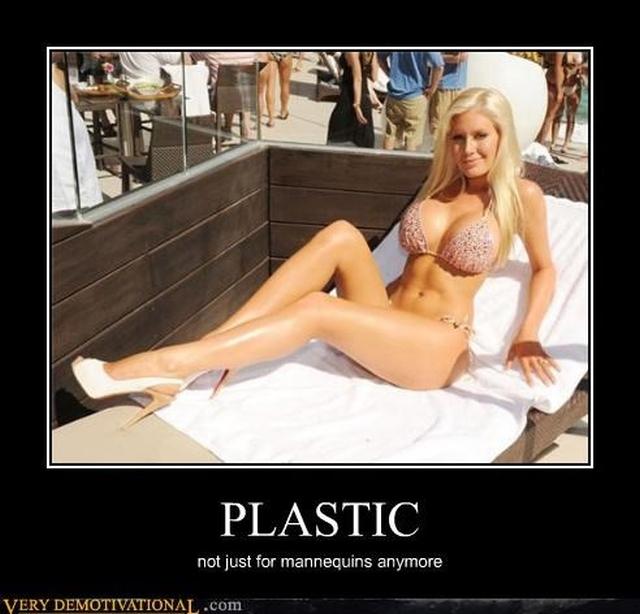}}
    \textit{\textbf{Label:} Misogynistic}
\end{minipage}
\hfill
\begin{minipage}[c]{0.65\textwidth}
    \begin{tcolorbox}[colback=white, colframe=black!10,
                      boxrule=0pt, sharp corners, 
                      valign=center,
                      left=2pt, right=2pt, top=2pt, bottom=2pt,
                      boxsep=0pt, enhanced]
        \begin{tcolorbox}[colback=orange!80!black!10!white, colframe=orange!80!black,
                          title=Prompt A Output (Surface-level description), fonttitle=\bfseries,
                          boxrule=0.8pt, sharp corners, fontupper=\small,
                          left=5pt, right=5pt, top=5pt, bottom=5pt, rounded corners,]
        The image shows a person lying on a lounge chair, wearing a bikini top and high-heeled shoes. The background appears to be an outdoor setting with people in the distance. The text at the bottom of the image reads "PLASTIC not just for mannequins anymore".
        \end{tcolorbox}

        \begin{tcolorbox}[colback=customblue2!10!white, colframe=customblue2,
                          title=Prompt B Output (Multimodal semantic inference), fonttitle=\bfseries,
                          boxrule=0.8pt, sharp corners, fontupper=\small,
                          left=5pt, right=5pt, top=5pt, bottom=5pt, rounded corners,]
        The meme uses a \textbf{provocative} image and text to make a statement about plastic surgery, implying that it is now common among women. The use of the word "plastic" in this context can be seen as a commentary on the perceived \textbf{superficiality and artificiality} of modern beauty standards. However, the meme does not explicitly criticize or promote any particular behavior but rather presents a \textbf{satirical view on societal norms} regarding appearance.
        \end{tcolorbox}
    \end{tcolorbox}
\end{minipage}

\caption{\textbf{Illustrative MAMI examples showing the outputs of our prompt-based LLM captioning strategies.} \textit{Left:} meme image. \textit{Right:} corresponding surface-level and deep-inference captions (Prompt A and Prompt B).}
\label{fig:llm_caption_example_mami}
\end{figure*}

\begin{figure*}[!t]
\centering
\begin{minipage}[c]{0.3\textwidth}
    \centering
    \fbox{\includegraphics[width=\linewidth,height=0.45\textheight,keepaspectratio]{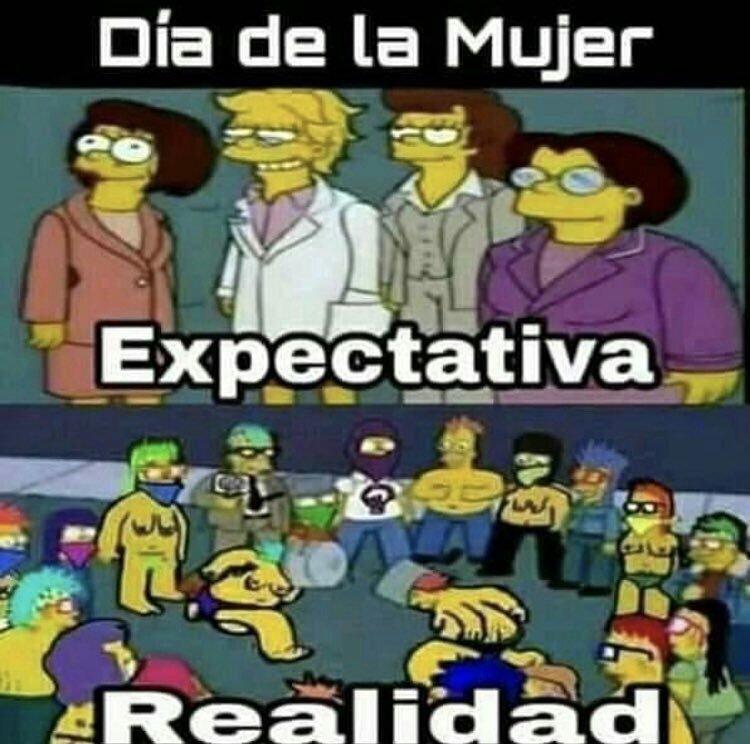}}
    \textit{\textbf{Label:} Sexist}
\end{minipage}
\hfill
\begin{minipage}[c]{0.65\textwidth}
    \begin{tcolorbox}[colback=white, colframe=black!10,
                      boxrule=0pt, sharp corners, 
                      valign=center, 
                      left=2pt, right=2pt, top=2pt, bottom=2pt,
                      boxsep=0pt, enhanced]
        \begin{tcolorbox}[colback=orange!80!black!10!white, colframe=orange!80!black,
                          title=Prompt A Output (Surface-level description), fonttitle=\bfseries,
                          boxrule=0.8pt, sharp corners, fontupper=\small,
                          left=5pt, right=5pt, top=5pt, bottom=5pt, rounded corners,]
        The image is a meme featuring characters from the animated TV show "The Simpsons." The top half of the image shows four adult female characters standing together, labeled "Expectativa" (Expectation), while the bottom half depicts a chaotic scene with various characters in costumes and masks, labeled "Realidad" (Reality). The contrast between the two halves \textbf{humorously} illustrates the difference between expectations and reality, particularly in the context of the Women's Day celebration.
        \end{tcolorbox}

        \begin{tcolorbox}[colback=customblue2!10!white, colframe=customblue2,
                          title=Prompt B Output (Multimodal semantic inference), fonttitle=\bfseries,
                          boxrule=0.8pt, sharp corners, fontupper=\small,
                          left=5pt, right=5pt, top=5pt, bottom=5pt, rounded corners,]
        The meme uses characters from "The Simpsons" to depict the "Expectativa" (expectation) and "Realidad" (reality) of the "Día de la Mujer" (International Women's Day). The top part shows four women in professional attire, representing the idealized expectation for International Women's Day. The bottom part shows a chaotic scene with various characters in costumes, suggesting the reality of the day. [...] This comparison
        \textbf{might be interpreted as a critique} of how International Women's Day is sometimes celebrated or perceived. However, this interpretation is subjective and depends on the viewer's perspective. In general, the meme does \textbf{not contain overtly sexist content} but rather uses humor to highlight a contrast. It's important to note that humor can be culturally specific and may not be universally understood or appreciated.
        \end{tcolorbox}
    \end{tcolorbox}
\end{minipage}%
\vspace{0.5cm}
\hrule
\vspace{0.5cm}
\begin{minipage}[c]{0.3\textwidth}
    \centering
    \fbox{\includegraphics[width=\linewidth,height=0.45\textheight,keepaspectratio]{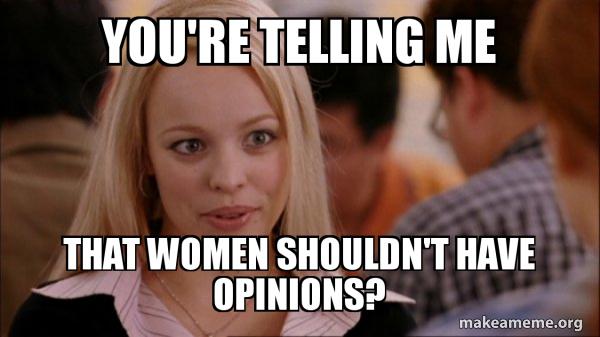}}
    \textit{\textbf{Label:} Non-Sexist}
\end{minipage}
\hfill
\begin{minipage}[c]{0.65\textwidth}
    \begin{tcolorbox}[colback=white, colframe=black!10,
                      boxrule=0pt, sharp corners, 
                      valign=center,
                      left=2pt, right=2pt, top=2pt, bottom=2pt,
                      boxsep=0pt, enhanced]
        \begin{tcolorbox}[colback=orange!80!black!10!white, colframe=orange!80!black,
                          title=Prompt A Output (Surface-level description), fonttitle=\bfseries,
                          boxrule=0.8pt, sharp corners, fontupper=\small,
                          left=5pt, right=5pt, top=5pt, bottom=5pt, rounded corners,]
        The image shows a person with blonde hair looking surprised or confused, with another person partially visible in the background. The setting appears to be indoors, possibly a social gathering or meeting.
        \end{tcolorbox}
        \begin{tcolorbox}[colback=customblue2!10!white, colframe=customblue2,
                          title=Prompt B Output (Multimodal semantic inference), fonttitle=\bfseries,
                          boxrule=0.8pt, sharp corners, fontupper=\small,
                          left=5pt, right=5pt, top=5pt, bottom=5pt, rounded corners,]
        The meme
        could be interpreted as \textbf{promoting gender stereotypes}, specifically suggesting that women should not have opinions. \textbf{This is a form of sexism} as it implies that women's perspectives and thoughts are less valid or important than those of men. The meme
        expresses disbelief or frustration, which can be seen as a criticism of the idea that women should be limited in their ability to have opinions. Therefore, the meme \textbf{does contain sexist content} by implicitly promoting a gender-based limitation on opinion formation.
        \end{tcolorbox}
    \end{tcolorbox}
\end{minipage}%
\vspace{0.5cm}
\hrule
\vspace{0.5cm}
\begin{minipage}[c]{0.3\textwidth}
    \centering
    \fbox{\includegraphics[width=\linewidth,height=0.45\textheight,keepaspectratio]{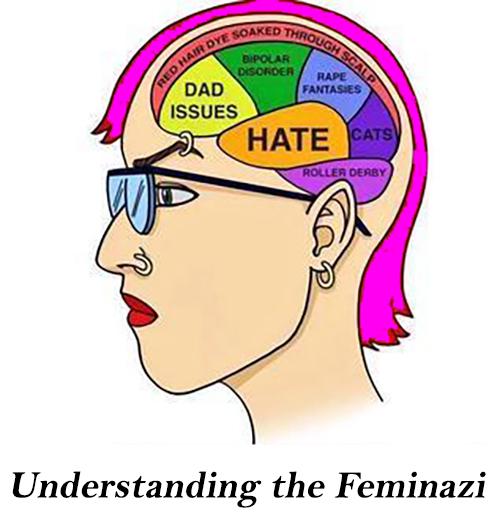}}
    \textit{\textbf{Label:} Sexist}
\end{minipage}
\hfill
\begin{minipage}[c]{0.65\textwidth}
    \begin{tcolorbox}[colback=white, colframe=black!10,
                      boxrule=0pt, sharp corners, 
                      valign=center,
                      left=2pt, right=2pt, top=2pt, bottom=2pt,
                      boxsep=0pt, enhanced]
        \begin{tcolorbox}[colback=orange!80!black!10!white, colframe=orange!80!black,
                          title=Prompt A Output (Surface-level description), fonttitle=\bfseries,
                          boxrule=0.8pt, sharp corners, fontupper=\small,
                          left=5pt, right=5pt, top=5pt, bottom=5pt, rounded corners,]
        The image shows a person lying on a lounge chair, wearing a bikini top and high-heeled shoes. The background appears to be an outdoor setting with people in the distance. The text at the bottom of the image reads "PLASTIC not just for mannequins anymore".
        \end{tcolorbox}

        \begin{tcolorbox}[colback=customblue2!10!white, colframe=customblue2,
                          title=Prompt B Output (Multimodal semantic inference), fonttitle=\bfseries,
                          boxrule=0.8pt, sharp corners, fontupper=\small,
                          left=5pt, right=5pt, top=5pt, bottom=5pt, rounded corners,]
        The meme uses a \textbf{provocative} image and text to make a statement about plastic surgery, implying that it is now common among women. The use of the word "plastic" in this context can be seen as a commentary on the perceived \textbf{superficiality and artificiality} of modern beauty standards. However, the meme does not explicitly criticize or promote any particular behavior but rather presents a \textbf{satirical view on societal norms} regarding appearance.
        \end{tcolorbox}
    \end{tcolorbox}
\end{minipage}
\caption{\textbf{Illustrative EXIST examples showing the outputs of our prompt-based LLM captioning strategies.} \textit{Left:} meme image. \textit{Right:} corresponding surface-level and deep-inference captions (Prompt A and Prompt B).}
\label{fig:llm_caption_example_exist}
\end{figure*}

Figures~\ref{fig:llm_caption_example_mami} and~\ref{fig:llm_caption_example_exist} provide examples of our LLM-based captioning strategies, contrasting the surface-level description (Prompt A) with the deeper semantic inference (Prompt B).

\end{document}